# Solving Limited Memory Influence Diagrams*

Denis Deratani Mauá denis@idsia.ch
Cassio Polpo de Campos cassio@idsia.ch
Marco Zaffalon zaffalon@idsia.ch

Istituto Dalle Molle di Studi sull'Intelligenza Articiale (IDSIA)
Galleria 2, Manno, 6928 - Switzerland

September 23, 2018

**Abstract**

We present a new algorithm for exactly solving decision making problems represented as influence diagrams. We do not require the usual assumptions of no forgetting and regularity; this allows us to solve problems with simultaneous decisions and limited information. The algorithm is empirically shown to outperform a state-of-the-art algorithm on randomly generated problems of up to 150 variables and $10^{64}$ solutions. We show that the problem is NP-hard even if the underlying graph structure of the problem has small treewidth and the variables take on a bounded number of states, but that a fully polynomial time approximation scheme exists for these cases. Moreover, we show that the bound on the number of states is a necessary condition for any efficient approximation scheme.

## 1 Introduction

Influence diagrams [12] are graphical models for utility-based decision making under uncertainty. Traditionally, they are designed to represent and solve situations involving a single, non-forgetful decision maker. Limited memory influence diagrams (hereafter LIMIDs) are generalizations of influence diagrams that allow for multi-agent and limited information decision problems [16, 23].[1] More precisely, LIMIDs relax the *regularity* and *no forgetting* assumptions of influence diagrams, namely, that there is a complete temporal ordering over the decision variables, and that at any step any previously disclosed information is remembered and taken into account to make a decision. These assumptions fail when decisions can be made simultaneously (e.g., by non-interacting agents), or when we wish to limit the decision history for computational reasons (e.g., to avoid an exponential blow up in the size of policies).

[1]Historically, Howard and Matheson [12] defined influence diagrams as graphical representations of general decision scenarios, and referred to the special cases respecting regularity and no forgetting as *decision networks*. The latter was then used by Zhang et al. [23] to describe the general case. Here, we adopt the more recent terminology of Lauritzen and Nilsson [16].

Solving a (limited memory) influence diagram refers to finding a combination of local decision rules (called a strategy) that maximizes expected utility. This task has been empirically and theoretically shown to be very hard [5]. In fact, we show here that solving a LIMID is NP-complete if we admit only singly connected diagrams with number of states per variable no greater than three. In addition, we show that an algorithm that produces provably good approximations within any fixed factor is unlike to exist even for diagrams with low treewidth, but that a fully polynomial time approximation scheme exists if we further restrict the variables to take on a bounded number of states.

Under certain graph-structural conditions (which no forgetting and regularity imply), Lauritzen and Nilsson [16] show that LIMIDs can be solved by a dynamic programming procedure with complexity exponential in the treewidth. However, when these conditions are not met, their iterative algorithm is only guaranteed to converge to a local optimum. Recently, de Campos and Ji [5] formulated the CR algorithm that maps a LIMID into a credal network [3] and then solves the corresponding marginal inference problem by mixed integer linear programming. By that, they were able to solve small problems exactly and obtain good approximations for medium-sized problems.

In this paper, we show that (partial) combinations of local decision rules can be partially ordered according to the utility they induce, and that dominance of a partial combination implies dominance of all (full) combinations that extend it. This greatly reduces the search space of strategies. Using these results we develop a generalized variable elimination procedure that computes the optimal solution by propagating only non-dominated (partial) solutions. We show experimentally that the algorithm can enormously save computational resources, and compute exact solutions for medium-sized problems. In fact, the algorithm is orders of magnitude faster than the CR algorithm on randomly generated diagrams containing up to 150 variables and $10^{64}$ strategies.

The paper is organized as follows. Section 2 formally describes LIMIDs and presents new results about the complexity of solving a LIMID. In Section 3, we present a new algorithm for computing exact global solutions and discuss its complexity. Section 4 contains the modifications necessary to convert the algorithm into a fully polynomial time approximation scheme for diagrams of bounded treewidth and number of states per variable. The performance of the algorithms is evaluated in Section 5. Finally, Sections 6 and 7 contain related work and final discussion. To improve readability, some of the proofs and supporting results appear in the appendix.

# 2 Limited Memory Influence Diagrams

In the formalism of (limited memory) influence diagrams, the quantities and events of interest are represented by three distinct types of variables or nodes.[2] *Chance variables* represent events on which the decision maker has no control, such as outcomes of tests or consequences of actions. *Decision variables* represent the options a decision maker might have. Finally, *value variables* represent additive parcels of the utility associated to a state of the world. The set of all variables considered relevant for a problem is denoted by $\mathcal{U}$. Each variable

[2] We make no distinction between a node in the graphical representation of a decision problem and its corresponding variable.

$X$ in $\mathcal{U}$ has an associated *domain* $\Omega_X$, which is the finite non-empty set of values $X$ can take on. The elements of $\Omega_X$ are called *states*. We assume the existence of the *empty domain* $\Omega_\emptyset \triangleq \{\lambda\}$, which contains a single element $\lambda$ which is not in any other domain. Decision and chance variables are assumed to have domains different from the empty domain, whereas value variables are always associated to the empty domain. The domain $\Omega_x$ of a set of variables $x = \{X_1, \dots, X_n\} \subseteq \mathcal{U}$ is given by the Cartesian product $\Omega_{X_1} \times \cdots \times \Omega_{X_n}$ of the variable domains. Thus, an element $\boldsymbol{u} \in \Omega_\mathcal{U}$ defines a state of the world, that is, a realization of all actions and events of interest. If $x$ and $y$ are sets of variables such that $y \subseteq x \subseteq \mathcal{U}$, and $\boldsymbol{x}$ is an element of the domain $\Omega_x$, we write $\boldsymbol{x}^{\downarrow y}$ to denote the projection of $\boldsymbol{x}$ onto the smaller domain $\Omega_y$, that is, $\boldsymbol{x}^{\downarrow y} \in \Omega_y$ contains only the components of $\boldsymbol{x}$ that are compatible with the variables in $y$. By convention, $\boldsymbol{x}^{\downarrow \emptyset} \triangleq \lambda$. The *cylindrical extension* of $\boldsymbol{y} \in \Omega_y$ to $\Omega_x$ is the set $\boldsymbol{y}^{\uparrow x} \triangleq \{\boldsymbol{x} \in \Omega_x : \boldsymbol{x}^{\downarrow y} = \boldsymbol{y}\}$. Often, we write $X_1 \cdots X_n$ to denote the set $\{X_1, \dots, X_n\}$ and, if clear from the context, $X$ to denote the singleton $\{X\}$.

Some operations over real-valued functions need to be defined. Let $f$ and $g$ be functions over domains $\Omega_x$ and $\Omega_y$, respectively. The product $fg$ is defined as the function over domain $\Omega_{x \cup y}$ such that $(fg)(\boldsymbol{w}) = f(\boldsymbol{w}^{\downarrow x})g(\boldsymbol{w}^{\downarrow y})$ for any $\boldsymbol{w}$ of its domain. Sum of functions is defined analogously: $(f+g)(\boldsymbol{w}) = f(\boldsymbol{w}^{\downarrow x})+g(\boldsymbol{w}^{\downarrow y})$. Notice that product and sum of functions are associative and commutative, and that product distributes over sum, that is, $fg = gf$, $f + g = g + f$, and $f(g+h) = fg+fh$. If $f$ is a function over $\Omega_x$, and $y \subseteq \mathcal{U}$, the *sum-marginal* $\sum_y f$ returns a function over $\Omega_{x \setminus y}$ such that for any element $\boldsymbol{w}$ of its domain we have $(\sum_y f)(\boldsymbol{w}) = \sum_{\boldsymbol{x} \in \boldsymbol{w}^{\uparrow x}} f(\boldsymbol{x})$. Notice that if $y \cap x = \emptyset$, then $\sum_y f = f$. Also, the sum-marginal operation inherits commutativity and associativity from addition of real numbers, and hence $\sum_{x \cup y} f = \sum_{x \setminus y} \sum_y f = \sum_{y \setminus x} \sum_x f$. If $\{f_x^{\boldsymbol{y}}\}_{\boldsymbol{y} \in \Omega_y}$ is a set containing functions $f_x^{\boldsymbol{y}}$ with domain $\Omega_x$, one for each element of $\Omega_y$, we write $f_x^y$ to denote the function that for all $\boldsymbol{w} \in \Omega_{x \cup y}$ satisfies $f_x^y(\boldsymbol{w}) = f_x^{\boldsymbol{w}^{\downarrow y}}(\boldsymbol{w}^{\downarrow x})$. For instance, if $X$ and $Y$ are two binary-valued variables, and $f_X^{\boldsymbol{y_1}} = (f_X^{\boldsymbol{y_1}}(\boldsymbol{x_1}), f_X^{\boldsymbol{y_1}}(\boldsymbol{x_2})) = (1/2, 1/2)$ and $f_X^{\boldsymbol{y_2}} = (0, 1)$ two functions over $\{X\}$, then the function $f_X^Y = (f_X^{\boldsymbol{y_1}}(\boldsymbol{x_1}), f_X^{\boldsymbol{y_1}}(\boldsymbol{x_2}), f_X^{\boldsymbol{y_2}}(\boldsymbol{x_1}), f_X^{\boldsymbol{y_2}}(\boldsymbol{x_2})) = (1/2, 1/2, 0, 1)$. If clear from the context, we write 1 to denote a function that returns one to all values in its domain and 0 to denote a function that returns always zero. More general, we write $k$ to denote a function that returns always a constant real value $k$. Hence, if $f$ and $g$ are functions over a domain $\Omega_x$ and $k$ is a real number, the expressions $f \geq g$ and $f = k$ denote that $f(\boldsymbol{x}) \geq g(\boldsymbol{x})$ and $f(\boldsymbol{x}) = k$, respectively, for all $\boldsymbol{x} \in \Omega_x$. Finally, any function over a domain containing a single element is identified with the real number it returns.

Let $\mathcal{C}$ denote the set of chance variables in $\mathcal{U}$, $\mathcal{D}$ the set of decision variables, and $\mathcal{V}$ the set of value variables. The sets $\mathcal{C}$, $\mathcal{D}$ and $\mathcal{V}$ form a partition of $\mathcal{U}$. A LIMID $\mathcal{L}$ consists of a direct acyclic graph (DAG) over the set of variables $\mathcal{U}$ annotated with variable types (decision, chance and value), together with a collection of (conditional) probability mass functions (one for each chance value) and utility functions (one for each value variable).[3] The value nodes in the graph have no children. The precise meanings of the arcs in $\mathcal{L}$ vary according to the type of node to which they point. Arcs entering chance and value nodes denote stochastic and functional dependency, respectively; arcs entering decision nodes describe information awareness or relevance at the time the decision is made. If $X$ is a node in $\mathcal{L}$, we denote by $\mathtt{pa}_X$ the set of parents of $X$, that is, the set of nodes of $\mathcal{L}$ from which there is an arc pointing to $X$. Similarly, we let $\mathtt{ch}_X$ denote the set of children of $X$ (i.e., nodes to which there is an arc from $X$), and $\mathtt{fa}_X \triangleq \mathtt{pa}_X \cup \{X\}$ denote its family. The descendants of $X$ are all nodes to which there is a directed path from $X$ in $\mathcal{L}$. Each chance variable $C$ in $\mathcal{C}$ has an associated set $\{p_C^{\boldsymbol{\pi}} : \boldsymbol{\pi} \in \Omega_{\mathtt{pa}_C}\}$ of (conditional) probability mass functions $p_C^{\boldsymbol{\pi}}$ quantifying the decision maker's beliefs about states $\boldsymbol{x} \in \Omega_C$ conditional on a state $\boldsymbol{\pi}$ of its parents (if $C$ has no parents, it has a single probability mass function assigned). We assume any chance variable $X \in \mathcal{C}$ to be stochastically independent from its non-descendant non-parents given its parents. Each value variable $V \in \mathcal{V}$ is associated with a real-valued utility function $u_V$ over $\Omega_{\mathtt{pa}_V}$, which quantifies the (additive) contribution of the states of its parents to the overall utility. Thus, the overall utility of a state $\boldsymbol{x} \in \Omega_{\mathcal{C}\cup\mathcal{D}}$ is given by the sum of utility functions $\sum_{V \in \mathcal{V}} u_V(\boldsymbol{x}^{\downarrow \mathtt{pa}_V})$.

[3]When no confusion arises, we write $\mathcal{L}$ to refer both to the LIMID and to its directed graph.

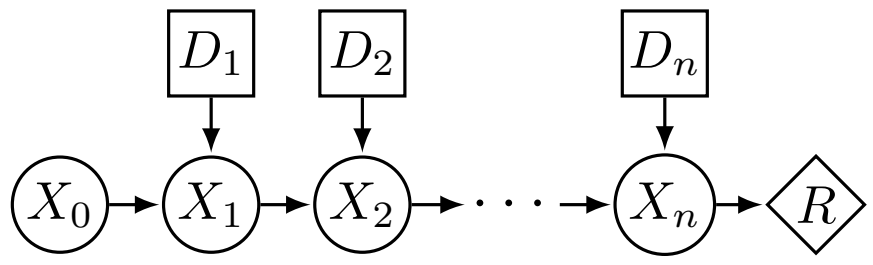

Figure 1: LIMID of the problem in the Example 1.

For any decision variable $D \in \mathcal{D}$, a *policy* $\delta_D$ specifies an action for each possible state configuration of its parents, that is, $\delta_D : \Omega_{\mathtt{pa}_D} \to \Omega_D$. If $D$ has no parents, then $\delta_D$ is a function from the empty domain to $\Omega_D$, and therefore constitutes a choice of $\boldsymbol{x} \in \Omega_D$. The set of all policies $\delta_D$ for a variable $D$ is denoted by $\Delta_D$.

The following artificial examples help to illustrate the use of LIMIDs for modeling decision-making problems under uncertainty involving multiple decision makers with limited information.

**Example 1.** *Consider a game where each of the $n$ participants has to decide between adding a ball to an urn or removing a ball from it, without knowing neither the state of the urn nor the other participants' decisions. If a participant decides for removal when the urn is empty then two balls are put. If the urn already contains two balls, a decision of adding a ball is ignored. The goal is to finish the sequence of $n$ decisions with no balls in the urn. To avoid intercommunication, the participants are kept in separate rooms, and they are asked for a decision in some pre-determined ordering unknown to them. Figure 1 depicts the graph structure of a LIMID modeling the problem. As usual, we graphically represent decision, chance, and value variables by squares, ovals and diamonds, respectively. The nodes $D_1, \ldots, D_n$ represent the decisions available to each of the participants in the given ordering. Each chance node $X_i$ represents the number of balls in the urn after $i$ decisions have been made (0,1, or 2), and is therefore associated to a deterministic function ($X_0$ models the initial state of the urn). The value node $R$ has an associated utility function that returns* 1 *if the state of $X_n$ is the element corresponding to an empty urn, and zero otherwise. For any $D_i$, a policy $\delta_{D_i}$ corresponds to a fixed decision of adding or removing a ball for the ith participant.*

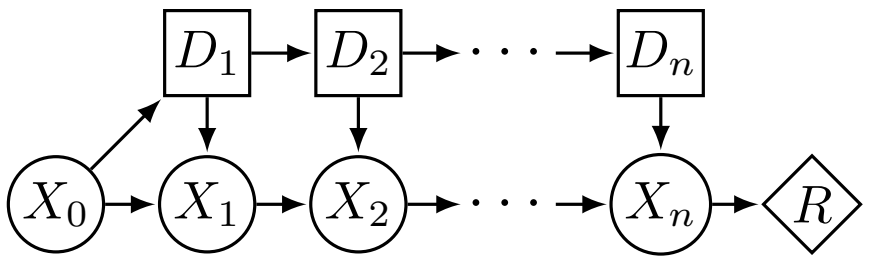

Figure 2: LIMID of the problem in the Example 2.

**Example 2.** *Consider a slightly modified version of the game in Example 1, where each participant is informed of the previous decision. Also, the initial state of the urn is disclosed to the first participant (so the first participant is aware he/she is the first to make a decision, but all the other remain ignorant about their positions in the sequence). Figure 2 depicts the graph structure of a LIMID modeling the problem. Notice the extra arcs from* $D_i$ *to* $D_{i-1}$, $i = 1, \ldots, n-1$, *representing that the decision made by the* $(i-1)$*-th participant is known to the* $i$*-th participant, and the arc* $X_0 \to D_1$ *indicating the known state of the urn to the first participant to make a decision. The policy* $\delta_{D_1}$ *prescribes a decision (to add or remove a ball) for the first participant for each possible initial state of the urn (empty, one ball or two balls). Similarly, for* $i = 2, \ldots, n$, *a policy* $\delta_{D_i}$ *specifies whether the* $i$*-th participant should decide to add a ball depending on the* $(i-1)$*-th participant's decision. For example, if* $\delta_{D_2} = (\delta_{D_2}(add), \delta_{D_2}(remove))$ *denotes the policy of the second participant (which is a function of the first participant's decision), then policy* $\delta_{D_2} = (remove, add)$ *prescribes that the second participant should remove a ball when the first participant decides for addition, and otherwise add a ball. The four possible policies for the second participant are* $\Delta_{D_2} = \{(add, remove), (add, add), (remove, add), (remove, remove)\}$.

Let $\Delta \triangleq \times_{D \in \mathcal{D}} \Delta_D$ denote the space of possible combination of policies. An element $s = (\delta_D)_{D \in \mathcal{D}} \in \Delta$ is said to be a *strategy* for $\mathcal{L}$. Given a policy $\delta_D$ and a state $\boldsymbol{\pi} \in \Omega_{\mathtt{pa}_D}$, let $p_D^{\boldsymbol{\pi}}$ denote a probability mass function for $D$ conditional on $\mathtt{pa}_D = \boldsymbol{\pi}$ such that $p_D^{\boldsymbol{\pi}}(\boldsymbol{x}) = 1$ if $\boldsymbol{x} = \delta_D(\boldsymbol{\pi})$ and $p_D^{\boldsymbol{\pi}}(\boldsymbol{x}) = 0$ otherwise. Hence, there is a one-to-one correspondence between functions $p_D^{\mathtt{pa}_D}$ and policies $\delta_D \in \Delta_D$, and specifying a policy $\delta_D$ is equivalent to specifying $p_D^{\mathtt{pa}_D}$. We denote the set of all functions $p_D^{\mathtt{pa}_D}$ obtained in this way by $\mathcal{P}_D$. A strategy $s$ induces a joint probability mass function over the variables in $\mathcal{C} \cup \mathcal{D}$ by

$$p_s \triangleq \prod_{C \in \mathcal{C}} p_C^{\mathtt{pa}_C} \prod_{D \in \mathcal{D}} p_D^{\mathtt{pa}_D}\,, \tag{1}$$

and has an associated expected utility given by

$$\mathrm{E}_s[\mathcal{L}] \triangleq \sum_{\boldsymbol{x} \in \Omega_{\mathcal{C} \cup \mathcal{D}}} p_s(\boldsymbol{x}) \sum_{V \in \mathcal{V}} u_V(\boldsymbol{x}^{\downarrow \mathtt{pa}_V}) \tag{2}$$

$$= \sum_{\mathcal{C} \cup \mathcal{D}} p_s \sum_{V \in \mathcal{V}} u_V\,. \tag{3}$$

Notice that the two sums in Eq. (3) have different semantics. The outer (leftmost) sum denotes the sum-marginal of the set of variables $\mathcal{C} \cup \mathcal{D}$, whereas the inner (rightmost) denotes the overall utility function over $\bigcup_{V \in \mathcal{V}} \mathtt{pa}_V$ that results from the sum of functions $u_V$.

The *treewidth* of a graph measures its resemblance to a tree and is given by the number of vertices in the largest clique of the corresponding triangulated moral graph minus one [1]. Given a LIMID $\mathcal{L}$ of treewidth $\omega$, we can evaluate the expected utility of any strategy $s$ in time and space at most exponential in $\omega$. Hence, if $\omega$ is bounded by a constant, obtaining $\mathrm{E}_s[\mathcal{L}]$ takes polynomial time:

**Proposition 3.** *Given a LIMID $\mathcal{L}$ with bounded treewidth and a strategy $s$, $\mathrm{E}_s[\mathcal{L}]$ can be computed in polynomial time.*

*Proof.* Given a strategy, the LIMID can be mapped into a LIMID with no decision nodes in polynomial time by replacing each decision node $D$ by a chance node with associated probability table $p_D^{\mathtt{pa}_D}$. The bounded treewidth implies that the number of parents of any decision node is bounded, and then encoding $p_D^{\mathtt{pa}_D}$ takes time polynomial in the number of elements in $\mathtt{fa}_D$ (an input of the LIMID). Then the LVE algorithm in Section 3 can be employed, which runs in polynomial time since there are no decisions in the input (therefore all sets are singletons) and the treewidth is bounded.[4] □

The primary task of a LIMID is to find a strategy $s^*$ with maximal expected utility, that is, to find $s^* \in \Delta$ such that

$$\mathrm{E}_s[\mathcal{L}] \leq \mathrm{E}_{s^*}[\mathcal{L}] \quad \text{for all } s. \tag{4}$$

The value $\mathrm{E}_{s^*}[\mathcal{L}]$ is called the *maximum expected utility* of $\mathcal{L}$ and it is denoted by $\mathrm{MEU}[\mathcal{L}]$. For most real problems, enumerating all the strategies is prohibitively costly. In fact, computing the MEU in bounded treewidth diagrams is NP-complete [5], and, as the following result implies, it remains NP-complete in even simpler LIMIDs.

**Theorem 4.** *Given a singly connected[5] LIMID with treewidth equal to 2, and with variables having at most three states, deciding whether there is a strategy with expected utility greater than a given $k$ is NP-complete.*

The proof, based on a reduction from the partition problem, is given in the appendix. Notice that the simple LIMIDs of Examples 1 and 2 (depicted in Figures 1 and 2, respectively) meet the preconditions of Theorem 4, and are in general most unlikely to be efficiently solvable for any sufficiently large $n$ (in fact, the hardness is shown by a reduction of an NP-complete problem to the problem of deciding whether the MEU of the LIMID in Example 1 is greater than a given threshold).

The complexity of solving a LIMID can be reduced by removing nodes and arcs that are irrelevant to the computation of the maximum expected utility. A chance or decision node is called *barren* if it has no children. Barren nodes have no influence on any value node and thus no impact on the MEU [8]. More irrelevances can be found by the concept of $d$-separation [19], which we succinctly state in the following paragraph.

A *trail* in $\mathcal{L}$ is a sequence of nodes such that any two consecutive nodes are connected by an arc. Notice that a trail does not need to "follow" the direction of the arcs. A trail $X, Z, Y$ is said to be *active* with respect to a set of variables $w$ either if $X$ and $Y$ are both parents of $Z$ and $Z$ or any of its descendants are in $w$, or if at least one of $X$ and $Y$ is not a parent of $Z$ and $Z$ is not in $w$. A trail is *blocked* by a set of variables $w$ if it contains a triple of consecutive nodes which is not active with respect to $w$. Two sets of nodes $x$ and $y$ are *d-separated* by a set of nodes $w$ if all trails from a node $X$ in $x$ to a node $Y$ in $y$ are blocked by $w$. Intuitively, if $X$ and $Y$ are $d$-separated by $w$, then $X$ and $Y$ are irrelevant to each other once we know the state of the variables in $w$.

[4] See also Koller and Friedman [15] for a slightly simpler variable elimination algorithm that computes fixed-strategy solutions of bounded treewidth diagrams in polynomial time.

[5] A directed graph is singly connected if the underlying (undirected) graph contains no cycles.

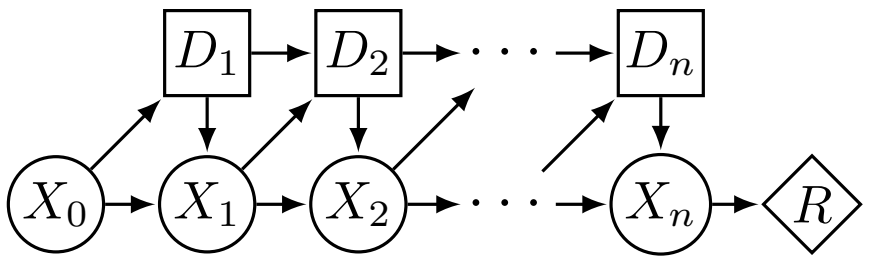

Figure 3: LIMID of the problem in the Example 5.

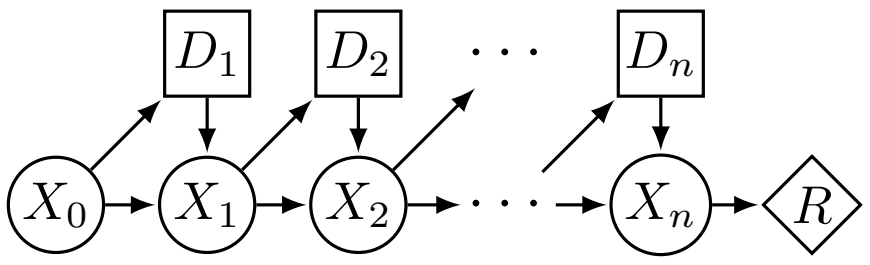

Figure 4: LIMID of the problem in the Example 5 after the removal of nonrequisite arcs.

A parent node $X$ of a decision node $D$ is *nonrequisite* (to $D$) if it is $d$-separated from all the value nodes that are descendant of $D$ given all the remaining parents and $D$. The arc from $X$ to $D$ is then said to be a *nonrequisite arc*. Likewise barren nodes, nonrequisite arcs can be removed without affecting the MEU [8, 16], and by doing so a node may become barren. We say that a LIMID is *minimal* if it contains no nonrequisite arcs and no barren nodes. Given a LIMID we can obtain its corresponding minimal diagram in polynomial time by repeatedly removing nonrequisite arcs and barren nodes [13, 16]. For the rest of this paper, we assume LIMIDs to be minimal.

The following example illustrates the concepts of $d$-separation and nonrequisite arcs.

**Example 5.** *Consider a modified version of the multiplayer game in Example 2, where in addition to the previous participant's decision also the current state of the urn is disclosed to a participant before he/she makes a decision. The graph structure of a LIMID representing this problem is depicted in Figure 3. Notice the arcs* $X_{i-1} \to D_i$*, for* $i = 1, \ldots, n-1$*, representing the known state of the urn to each participant. Consider the arc* $D_1 \to D_2$*. All trails connecting* $D_1$ *and* $R$ *are blocked by* $\mathtt{fa}_{D_2} \setminus \{D_1\} = \{D_2, X_1\}$*. Therefore* $D_1$ *is a nonrequisite node for* $D_2$ *and the arc* $D_1 \to D_2$ *can be removed without altering the MEU. In fact, all arcs* $D_{i-1} \to D_i$*, for* $i = 1, \ldots, n$*, are nonrequisite and can be safely removed. On the other hand,* $X_0$ *is a requisite node for* $D_1$ *because, for instance, the trail* $X_0, X_1, \ldots, X_n, R$ *is active given the set* $\mathtt{fa}_{D_1} \setminus \{X_0\} = \{D_1\}$*. Intuitively, the information provided by the arcs* $D_{i-1} \to D_i$ *(i.e., the previous participant's decision) does not help in making a decision once the current state* $X_{i-1}$ *of the* *urn is known. Figure 4 depicts a minimal version of the diagram in Figure 3.*

Under the usual assumptions of complexity theory, when a problem is NP-hard to solve the best available options are (i) trying to devise an algorithm that runs efficiently on many instances but has exponential worst-case complexity, or (ii) trying to develop an approximation algorithm that for all instances provides in polynomial time a solution that is provably within a certain range of the optimal solution. In Section 3, we take option (i), and present an algorithm that efficiently computes optimal solutions for many LIMIDs, but runs in exponential time for many others. Given $\epsilon > 0$, an $\epsilon$-approximation algorithm (for solving a LIMID) obtains a strategy $s$ such that $(1+\epsilon)\,\mathrm{E}_s[\mathcal{L}] \geq \mathrm{MEU}[\mathcal{L}]$. As the following result indicates, alternative (ii) is most likely unfeasible, even if we consider only diagrams of bounded treewidth.

**Theorem 6.** *Given a singly connected LIMID $\mathcal{L}$ with bounded treewidth, (unless P=NP) there is no polynomial time $\epsilon$-approximation algorithm, for any $0 < \epsilon < 2^\theta - 1$, where $\theta$ is the number of numerical parameters (probabilities and utilities) required to specify $\mathcal{L}$.*

We defer the proof to the appendix. However, we show in Section 4 that, differently from the general case, there are LIMIDs, namely, those with bounded treewidth and number of states per variable, for which a polynomial time $\epsilon$-approximation algorithm exists.

Let $\mathcal{L}$ be a LIMID and let $k$ and $K$ denote, respectively, the smallest and the greatest utilities associated to any of the value variables, that is, for all $V \in \mathcal{V}$ we have that $k \leq u_V \leq K$, and there are $V$ and $V'$ such that $u_V(\boldsymbol{x}) = k$ and $u_{V'}(\boldsymbol{x}') = K$ for some $\boldsymbol{x} \in \Omega_{\mathtt{pa}_V}$ and $\boldsymbol{x}' \in \Omega_{\mathtt{pa}_{V'}}$. Assume $k < K$ (otherwise the MEU is trivial), and let $\mathcal{L}'$ be the LIMID obtained from $\mathcal{L}$ by setting each utility function $u_V$ associated to a value node $V$ to $u'_V = (u_V - k)/(K-k)$. Note that by design each function $u'_V$ takes values on $[0,1]$. The following well-known result allows us to focus on scaled utility functions.

**Proposition 7.** *For any strategy $s$, $\mathrm{E}_s[\mathcal{L}] = (K-k)\,\mathrm{E}_s[\mathcal{L}'] + k|\mathcal{V}|$.*

*Proof.* The case of a single value node has been shown by Cooper [2] and Shachter and Peot [21]. The extension to multiple value nodes is straightforward. For any strategy $s$ we have that

$$\begin{aligned}
\mathrm{E}_s[\mathcal{L}'] &= \sum_{\mathcal{C}\cup\mathcal{D}} p_s \sum_{V\in\mathcal{V}} u'_V \\
&= \sum_{\mathcal{C}\cup\mathcal{D}} p_s \sum_{V\in\mathcal{V}} \frac{u_V - k}{K-k} \\
&= \frac{1}{K-k} \sum_{\mathcal{C}\cup\mathcal{D}} p_s \left( -k|\mathcal{V}| + \sum_{V\in\mathcal{V}} u_V \right) \\
&= \frac{1}{K-k} \left( -k|\mathcal{V}| \sum_{\mathcal{C}\cup\mathcal{D}} p_s + \sum_{\mathcal{C}\cup\mathcal{D}} p_s \sum_{V\in\mathcal{V}} u_V \right)
\end{aligned}$$

which, since $p_s$ is a probability distribution on $\mathcal{C} \cup \mathcal{D}$, equals

$$\frac{1}{K-k}\left(\mathrm{E}_s[\mathcal{L}] - k|\mathcal{V}|\right) .$$

Hence, the result follows. □

In the rest of the paper we consider only LIMIDs with utilities taking values in some subset of the real interval $[0, 1]$, which due to Proposition 7 does not incur any loss of generality.

# 3 Solving LIMIDs Exactly

The basic ingredients of our algorithmic framework for representing and handling information in LIMIDs are the so called *valuations*, which encode information (probabilities, utilities and policies) about the elements of a domain. Each valuation is associated to a subset of the variables in $\mathcal{U}$, called its *scope*. More concretely, a valuation $\phi$ with scope $x$ is a pair $(p, u)$ of bounded nonnegative real-valued functions $p$ and $u$ over the domain $\Omega_x$; we refer to $p$ and $u$ as the probability and utility part, respectively, of $\phi$. Often, we write $\phi_x$ to make explicit the scope $x$ of a valuation $\phi$. For any $x \subseteq \mathcal{U}$, we denoted the set of all possible valuations with scope $x$ by $\Phi_x$. The set of all possible valuations is thus given by $\Phi \triangleq \bigcup_{x\subseteq\mathcal{U}} \Phi_x$. The set $\Phi$ is closed under two basic operations of *combination* and *marginalization*. Combination represents the aggregation of information and is defined as follows.

**Definition 8.** *If $\phi = (p, u)$ and $\psi = (q, v)$ are valuations with scopes $x$ and $y$, respectively, its combination $\phi \otimes \psi$ is the valuation $(pq, pv + qu)$ with scope $x \cup y$.*

Marginalization, on the other hand, acts by coarsening information:

**Definition 9.** *If $\phi = (p, u)$ is a valuation with scope $x$, and $y$ is a set of variables such that $y \subseteq x$, the marginal $\phi^{\downarrow y}$ is the valuation $(\sum_{x\setminus y} p, \sum_{x\setminus y} u)$ with scope $y$. In this case, we say that $z \triangleq x \setminus y$ has been* eliminated *from $\phi$, which we denote by $\phi^{-z}$.*

Notice that our definitions of combination and marginalization differ from previous works on LIMIDs (e.g., [16]), which usually require a division of the utility part by the probability part. The removal of division turns out to be an important feature when we discuss maximality of valuations later on.

In terms of computational complexity, combining two valuations $\phi$ and $\psi$ with scopes $x$ and $y$, respectively, requires $3|\Omega_{x\cup y}|$ multiplications and $|\Omega_{x\cup y}|$ additions of numbers; computing $\phi^{\downarrow y}$, where $y \subseteq x$, costs $|\Omega_{x\cup y}|$ operations of addition. In other words, the cost of combining or marginalizing a valuation is exponential in the cardinality of its scope (and linear in the cardinality of its domain). Hence, we wish to work with valuations whose scope is as small as possible. The following result shows that our framework respects the necessary conditions for computing efficiently with valuations (in the sense of keeping the scope of valuations obtained from combinations and marginalizations of other valuations minimal).

**Proposition 10.** *The system $(\Phi, \mathcal{U}, \otimes, \downarrow)$ satisfies the following three axioms of a (weak)* labeled valuation algebra *[14, 22].*

*(A1) Combination is commutative and associative, that is, for any $\phi_1, \phi_2, \phi_3 \in \Phi$ we have that*

$$\phi_1 \otimes \phi_2 = \phi_2 \otimes \phi_1 \,,$$
$$\phi_1 \otimes (\phi_2 \otimes \phi_3) = (\phi_1 \otimes \phi_2) \otimes \phi_3 \,.$$

*(A2) Marginalization is transitive, that is, for $\phi_z \in \Phi_z$ and $y \subseteq x \subseteq z$ we have that*

$$(\phi_z^{\downarrow x})^{\downarrow y} = \phi_z^{\downarrow y} .$$

*(A3) Marginalization distributes over combination, that is, for $\phi_x \in \Phi_x$, $\phi_y \in \Phi_y$ and $x \subseteq z \subseteq x \cup y$ we have that*

$$(\phi_x \otimes \phi_y)^{\downarrow z} = \phi_x \otimes \phi_y^{\downarrow y \cap z} .$$

*Proof.* (A1) follows directly from commutativity, associativity and distributivity of product and sum of real-valued functions, and (A2) follows directly from commutativity of the sum-marginal operation. To show (A3), consider any two valuations $(p, u)$ and $(q, v)$ with scopes $x$ and $y$, respectively, and a set $z$ such that $x \subseteq z \subseteq x \cup y$. By definition of $\otimes$ and $\downarrow$, we have that

$$[(p, u) \otimes (q, v)]^{\downarrow z} = \left( \sum_{x \cup y \setminus z} pq, \sum_{x \cup y \setminus z} (pv + qu) \right) .$$

Since $x \cup y \setminus z = y \setminus z$, and $p$ and $u$ are functions over $\Omega_x$, it follows that

$$\begin{aligned} \left( \sum_{x \cup y \setminus z} pq, \sum_{x \cup y \setminus z} (pv + qu) \right) &= \left( p \sum_{y \setminus z} q, p \sum_{y \setminus z} v + u \sum_{y \setminus z} q \right) \\ &= (p, u) \otimes \left( \sum_{y \setminus z} q, \sum_{y \setminus z} v \right) , \end{aligned}$$

which equals $(p, y) \otimes (q, v)^{\downarrow y \cap z}$. □

The following is a direct consequence of (A3) that is required to prove the correctness of the variable elimination procedure.

**Lemma 11.** *If $\phi_x \in \Phi_x$, $\phi_y \in \Phi_y$, $z \subseteq y$ and $z \cap x = \emptyset$, then $(\phi_x \otimes \phi_y)^{-z} = \phi_x \otimes \phi_y^{-z}$.*

*Proof.* Let $w = x \cup y \setminus z$. Since $x \cap z = \emptyset$, it follows that $x \subseteq w \subseteq x \cup y$. Hence, by definition of elimination and (A3), we have that

$$\begin{aligned} (\phi_x \otimes \phi_y)^{-z} &= (\phi_x \otimes \phi_y)^{\downarrow w} \\ &= \phi_x \otimes \phi_y^{\downarrow y \cap w} . \end{aligned}$$

But $y \cap w = y \setminus z$. Thus, $\phi_y^{\downarrow y \cap w} = \phi_y^{-z}$. □

The following result shows how valuations can be used to compute expected utilities for a given strategy.

**Proposition 12.** *Given a LIMID $\mathcal{L}$ and a strategy $s = (\delta_D)_{D \in \mathcal{D}} \in \Delta$, let*

$$\phi_s \triangleq \left[ \bigotimes_{C \in \mathcal{C}} (p_C^{\mathsf{pa}_C}, 0) \right] \otimes \left[ \bigotimes_{D \in \mathcal{D}} (p_D^{\mathsf{pa}_D}, 0) \right] \otimes \left[ \bigotimes_{V \in \mathcal{V}} (1, u_V) \right] , \quad (5)$$

*where, for each $D$, $p_D^{\mathsf{pa}_D}$ is the function in $\mathcal{P}_D$ associated with policy $\delta_D$. Then $\phi_s^{\downarrow \emptyset}$ equals $(1, \mathrm{E}_s[\mathcal{L}])$.*

*Proof.* Let $p$ and $u$ denote the probability and utility part, respectively, of $\phi_s^{\downarrow\emptyset}$. By definition of combination, we have that $\phi_s = (p_s, p_s \sum_{V\in\mathcal{V}} u_V)$, where $p_s = \prod_{X\in\mathcal{C}\cup\mathcal{D}} p_X^{\mathrm{pa}_X}$ as in (1). Since $p_s$ is a probability distribution over $\mathcal{C}\cup\mathcal{D}$, it follows that $p = \sum_{\boldsymbol{x}\in\Omega_{\mathcal{C}\cup\mathcal{D}}} p_s(\boldsymbol{x}) = 1$. Finally, $u = \sum_{\mathcal{C}\cup\mathcal{D}} p_s \sum_{V\in\mathcal{V}} u_V$, which equals $\mathrm{E}_s[\mathcal{L}]$ by (3). □

Hence, given a strategy we can use a variable elimination procedure[6] to compute its expected utility in time polynomial in the largest domain of a variable but exponential in the width of the elimination ordering [e.g., 15, Section 23.4.3].[7] However, computing the MEU in this way is unfeasible for realistic diagrams due to the large number of strategies that would need to be enumerated. For example, a simple LIMID consisting of a decision variable with four chance nodes as parents and one value node as child contains $10^{3^4} = 10^{81}$ strategies in $\Delta$, if the decision variable has 10 states and each parent has 3 states.

In order to avoid having to consider all possible strategies, we define a partial order (i.e., a reflexive, antisymmetric and transitive relation) over $\Phi$ as follows.

**Definition 13.** *For any two valuations $\phi = (p, u)$ and $\psi = (q, v)$ in $\Phi$, we say that $\psi$* dominates *$\phi$ (conversely, we say that $\phi$* is dominated by *$\psi$), and we write $\phi \leq \psi$, if $\phi$ and $\psi$ have equal scope, $p \leq q$, and $u \leq v$.*

If $\phi$ and $\psi$ have scope $x$, deciding whether $\psi$ dominates $\phi$ costs at most $2|\Omega_x|$ operations of comparison of numbers. The following result shows that the algebra of valuations is monotonic with respect to dominance.

**Proposition 14.** *The system $(\Phi, \mathcal{U}, \otimes, \downarrow, \leq)$ satisfies the following two additional axioms of an* ordered valuation algebra *[11].*

*(A4) Combination is monotonic with respect to dominance, that is,*

$$\text{if } \phi_x \leq \psi_x \text{ and } \phi_y \leq \psi_y \text{ then } (\phi_x \otimes \phi_y) \leq (\psi_x \otimes \psi_y)\,.$$

*(A5) Marginalization is monotonic with respect to dominance, that is,*

$$\text{if } \phi_x \leq \psi_x \text{ then } \phi_x^{\downarrow y} \leq \psi_x^{\downarrow y}\,.$$

*Proof.* (A4). Consider two valuations $(p_x, u_x)$ and $(q_x, v_x)$ with scope $x$ such that $(p_x, u_x) \leq (q_x, v_x)$, and two valuations $(p_y, u_y)$ and $(q_y, v_y)$ with scope $y$ satisfying $(p_y, u_y) \leq (q_y, v_y)$. By definition of $\leq$, we have that $p_x \leq q_x$, $u_x \leq v_x$, $p_y \leq q_y$ and $u_y \leq v_y$. Since all functions are nonnegative, it follows that $p_x p_y \leq q_x q_y$, $p_x u_y \leq q_x v_y$ and $p_y u_x \leq q_y v_x$. Hence, $(p_x, u_x) \otimes (p_y, u_y) = (p_x p_y, p_x u_y + p_y u_x) \leq (q_x q_y, q_x v_y + q_y v_x) = (q_x, v_x) \otimes (q_y, v_y)$. (A5). Let $y$ be a subset of $x$. It follows from monotonicity of $\leq$ with respect to addition of real numbers that

$$(p_x, u_x)^{\downarrow y} = \left(\sum_{x\setminus y} p_x, \sum_{x\setminus y} u_x\right) \leq \left(\sum_{x\setminus y} q_x, \sum_{x\setminus y} v_x\right) = (q_x, v_x)^{\downarrow y}\,.$$

Hence, the result follows. □

[6] Variable elimination algorithms are also known in the literature as fusion algorithms [22] and bucket elimination [6].

[7] The width of an elimination ordering is the treewidth of the tree decomposition it induces and can be computed in time polynomial in the number of variables. It also equals the maximum cardinality of the scope of a valuation in the variable elimination procedure minus one.

The algorithm we devise later on operates on sets of ordered valuations.

**Definition 15.** *Given a finite set of valuations $\Psi \subseteq \Phi$, we say that $\phi \in \Psi$ is* maximal *if for all $\psi \in \Psi$ such that $\phi \leq \psi$ it holds that $\psi \leq \phi$. The operator* max *returns the set* $\max(\Psi)$ *of maximal valuations of* $\Psi$.

If $\Psi_x$ is a set with $m$ valuations with scope $x$, $\max(\Psi_x)$ can be obtained by $m^2$ comparisons $\phi \leq \psi$, where $(\phi, \psi) \in \Psi_x \times \Psi_x$.

We extend combination and marginalization to sets of valuations as follows.

**Definition 16.** *If $\Psi_x$ and $\Psi_y$ are any two sets of valuations in $\Phi$,*

$$\Psi_x \otimes \Psi_y \triangleq \{\phi_x \otimes \phi_y : \phi_x \in \Psi_x, \phi_y \in \Psi_y\}$$

*denotes the set obtained from all combinations of a valuation in $\Psi_x$ and a valuation in $\Psi_y$.*

**Definition 17.** *If $\Psi_x \subseteq \Phi_x$ is a set of valuations with scope $x$ and $y \subseteq x$,*

$$\Psi_x^{\downarrow y} \triangleq \{\phi_x^{\downarrow y} : \phi_x \in \Psi_x\}$$

*denote the set of valuations obtained by element-wise marginalization of valuations to $y$.*

It can be checked that sets of valuations with combination and marginalization defined element-wise satisfy axioms (A1)–(A3), and therefore form a valuation algebra. Hence, Lemma 11 applies also for sets of valuations with marginalization and combination defined as above.

**Lemma 18.** *If $\Psi_x \subseteq \Phi_x$ and $\Psi_y \subseteq \Phi_y$ are two sets of valuations with scope $x$ and $y$, respectively, and $z$ is a set of variables such that $z \subseteq y$ and $z \cap x = \emptyset$, then $(\Psi_x \otimes \Psi_y)^{-z} = \Psi_x \otimes \Psi_y^{-z}$.*

*Proof.* The result follows from element-wise application of Lemma 11 to $(\phi_x \otimes \phi_y)^{-z} \in (\Psi_x \otimes \Psi_y)^{-z}$. □

We are now ready to describe the **LVE algorithm**, which solves arbitrary LIMIDs exactly. Consider a LIMID $\mathcal{L}$, and an ordering $X_1 < \cdots < X_n$ over the variables in $\mathcal{C} \cup \mathcal{D}$. The algorithm is initialized by generating one set of valuations for each variable $X$ in $\mathcal{U}$ as follows.

**Initialization:** Let $\mathcal{V}_0$ be initially the empty set.

1. For each chance variable $X \in \mathcal{C}$, add a singleton $\Psi_X \triangleq \{(p_X^{\mathsf{pa}_X}, 0)\}$ to $\mathcal{V}_0$.
2. For each decision variable $X \in \mathcal{D}$, add a set of valuations $\Psi_X \triangleq \{(p_X^{\mathsf{pa}_X}, 0) : p_X^{\mathsf{pa}_X} \in \mathcal{P}_X\}$ to $\mathcal{V}_0$.
3. For each value variable $X \in \mathcal{V}$, add a singleton $\Psi_X \triangleq \{(1, u_X)\}$ to $\mathcal{V}_0$.

Once $\mathcal{V}_0$ has been initialized with a set of valuations for each variable in the diagram, we recursively eliminate a variable $X_i$ in $\mathcal{C} \cup \mathcal{D}$ in the given ordering and remove any non-maximal valuation:

**Propagation:** For $i = 1, \ldots, n$ do:

1. Let $\mathcal{B}_i = \emptyset$. Remove from $\mathcal{V}_{i-1}$ all sets whose valuations contain $X_i$ in their scope and add them to $\mathcal{B}_i$.

2. Compute $\Psi_i \triangleq \max([\bigotimes_{\Psi \in \mathcal{B}_i} \Psi]^{-X_i})$.

3. Set $\mathcal{V}_i \triangleq \mathcal{V}_{i-1} \cup \{\Psi_i\}$.

**Termination:** Finally, the algorithm outputs the utility part of the single maximal valuation in the set $\bigotimes_{\Psi \in \mathcal{V}_n} \Psi$, that is, the algorithm returns the real number $u$ such that $(p, u) \in \max(\bigotimes_{\Psi \in \mathcal{V}_n} \Psi)$. $u$ is a real number because the valuations in $\bigotimes_{\Psi \in \mathcal{V}_n} \Psi$ have empty scope and thus both their probability and utility parts can be identified with real numbers.

The elimination ordering $X_1 < \cdots < X_n$ can be determined using the standard heuristics for variable elimination in Bayesian networks such as minimizing the number of fill-ins or the cardinality of the domain of the neighbor set [13, 15].

Differently from other message-passing algorithms that obtain approximate solutions to LIMIDs by (repeatedly) propagating a single valuation (e.g., the SPU algorithm [16]), the LVE algorithm computes exact solutions by propagating many maximal valuations that correspond to partial combinations of local decision rules. The efficiency of the algorithm in handling the propagation of several valuations derives from the early removal of valuations performed by the max operation in the propagation step.

Consider the set $\Psi_{\mathcal{L}} \triangleq \{\phi_s : s \in \Delta\}$, where each $\phi_s$ is given by (5). It is not difficult to see that

$$
\begin{aligned}
\Psi_{\mathcal{L}} &= \left[\bigotimes_{C \in \mathcal{C}} \left\{\left(p_C^{\mathsf{pa}_C}, 0\right)\right\}\right] \otimes \left[\bigotimes_{D \in \mathcal{D}} \left\{\left(p_D^{\mathsf{pa}_D}, 0\right) : p_D^{\mathsf{pa}_D} \in \mathcal{P}_D\right\}\right] \otimes \left[\bigotimes_{V \in \mathcal{V}} \{(1, u_V)\}\right] \\
&= \bigotimes_{\Psi_X \in \mathcal{V}_0} \Psi_X \,.
\end{aligned}
$$

Hence, by Proposition 12 we have that each $\phi_s^{\downarrow\emptyset}$ in $\Psi_{\mathcal{L}}^{\downarrow\emptyset}$ is a valuation with probability part one and utility part equal to the expected utility of some strategy in $\Delta$. Since the relation $\leq$ induces a strict (linear) order over $\Psi_{\mathcal{L}}^{\downarrow\emptyset}$, the MEU of the diagram equals the utility part of the (single) valuation in $\max(\Psi_{\mathcal{L}}^{\downarrow\emptyset})$. The variable elimination procedure in the propagation step is responsible for obtaining $\max(\bigotimes_{\Psi \in \mathcal{V}_n} \Psi) = \max(\Psi_{\mathcal{L}}^{\downarrow\emptyset})$ more efficiently by distributing max and $\downarrow$ over $\bigotimes_{\Psi_X \in \mathcal{V}_0} \Psi_X$, which allows for a significant reduction in the cardinalities of sets and scopes of valuations produced. The following result states the correctness of the algorithm.

**Theorem 19.** *Given a LIMID $\mathcal{L}$, LVE outputs* $\text{MEU}[\mathcal{L}]$.

The proof, which requires some technicalities, is in the appendix.

## 3.1 Complexity Analysis

Assume that decision nodes are parentless. We will show in Section 3.4 later on that we can transform any given LIMID into an equivalent model in which decision nodes have no parents. Parentless decisions allows us to avoid having to deal with sets whose cardinality is exponential in the number of parents.

The time complexity of the algorithm is given by the cost of creating the sets of valuations in the initialization step plus the overall cost of the combination and marginalization operations performed during the propagation step. Regarding the initialization step, the loops for chance and value variables generate singletons, and thus take time linear in the input. Since decision nodes have no parents, there are $\rho_D \triangleq |\Omega_D|$ policies in $\Delta_D$ (which coincides with the number of functions in $\mathcal{P}_D$) for each decision variable $D$. There is one valuation in the corresponding set $\Psi_D$ added to $\mathcal{V}_0$ for every policy in $\Delta_D$. Let $\rho \triangleq \max_{D \in \mathcal{D}} \rho_D$ be the cardinality of the largest policy set. Then the initialization loop for decision variables takes $O(|\mathcal{D}|\rho)$ time, which is polynomial in the input. Let us now analyze the propagation step. As with any variable elimination procedure, the running time of propagating (sets of) valuations is exponential in the width of the elimination ordering, which is in the best case given by the treewidth of the diagram. Consider the case of an elimination ordering with bounded width $\omega$, and a diagram with bounded number of states per variable $\kappa$. Then the cost of each combination or marginalization is bounded by a constant, and the complexity depends only on the number of operations performed. Moreover, we have in this case that $\rho \leq \kappa$. Let $\nu$ denote the cardinality of the largest set $\Psi_i$, for $i = 1, \ldots, n$. Thus, computing $\Psi_i$ requires at most $\nu^{|\mathcal{U}|-1}$ operations of combination (because that is the maximum number of sets that we might need to combine to compute $\bigotimes_{\Psi \in \mathcal{B}_i} \Psi$ in the propagation step) and $\nu$ operations of marginalization. In the worst case, $\nu$ is equal to $\rho^{|\mathcal{D}|} \leq O(\kappa^{|\mathcal{D}|})$, that is, all sets associated to decision variables have been combined without discarding any valuation. Hence, the worst-case complexity of the propagation step is exponential in the number of decision variables, even if the width of the elimination ordering and the number of states per variable are bounded. Note however that this is a very pessimistic scenario and, on average, the removal of non-maximal elements greatly reduces the complexity, as the experiments in Section 5 show.

## 3.2 Strategy Selection

Most likely, one is not only interested in the maximum expected utility of a LIMID but also in an optimum course of action for every possible scenario, that is, in an optimal strategy that obtains the MEU. LVE can be easily modified to provide an optimal strategy by storing at each step the policies associated to non-dominated valuations as follows. For each valuation created in the initialization step associate a list which is empty unless the valuation refers to a policy of a decision variable, in which case the list contains the associated policy. Now, any valuation $\phi_i \in \Psi_i$ factorizes as $(\psi_1 \otimes \cdots \otimes \psi_{|\mathcal{B}_i|})^{-X_i}$, where each $\psi_j$ is an element of a different set $\Psi$ in $\mathcal{B}_i$. For $i = 1$, each $\psi_j$ is associated to a list. Assign a list to each $\phi_1$ which equals the concatenation of the lists of its factors $\psi_j$. Thus, the list contains a choice of policies for all decision variables $D$ such that $X_1 \in \mathtt{fa}_D$. For each $i$, associate a list to each $\phi_i \in \Psi_i$ which equals the concatenation of lists of its factors. An optimal strategy $s^*$ is thus easily obtained from the list associated to $\phi_n \in \max(\bigotimes_{\Psi \in \mathcal{V}_n} \Psi)$. The handling of lists can be implemented by simple pointers to valuations in sets in $\mathcal{V}_0$, and therefore the asymptotic complexity of the algorithm is unaltered.

## 3.3 Reverse Topological Ordering

The valuations used by LVE specify twice as many numbers as the cardinality of the domain of their associated scope. It is possible to decrease by a factor of two the number of numerical parameters per valuation the algorithm needs to handle by constraining the elimination of variables to follow a reverse topological ordering in the diagram, that is, by requiring each variable to be processed only after all its descendants have been processed. As the following result shows, any reverse topological ordering produces valuations whose probability part equals one in all coordinates.

**Proposition 20.** *If $X_1 < \cdots < X_n$ denotes a reverse topological ordering over the variables in $\mathcal{C} \cup \mathcal{D}$, then for $i = 1, \ldots, n$ the valuations in $\Psi_i$ have probability part $p = 1$, where $1$ is the function that always returns the unity.*

*Proof.* We show the result by induction on $i$. Regarding the basis, we have from the reverse topological ordering that $X_1$ is a variable containing only value nodes as children. Hence, $\mathcal{B}_1 = \{\Psi_{X_1}\} \cup \{\{(1, u_V)\} : V \in \mathtt{ch}_{X_1}\}$, where by definition $\Psi_{X_1}$ equals $\{(p_{X_1}^{\mathtt{pa}_{X_1}}, 0)\}$ if $X_1$ is a chance node, and $\{(p_{X_1}^{\mathtt{pa}_{X_1}}, 0) : p_{X_1}^{\mathtt{pa}_{X_1}} \in \mathcal{P}_{X_1}\}$ if it is a decision node. It follows that

$$\Psi_1 = \max\left(\left\{\Big(\sum_{X_1} p_{X_1}^{\mathtt{pa}_{X_1}}, \sum_{X_1} p_{X_1}^{\mathtt{pa}_{X_1}} \sum_{V \in \mathtt{ch}_{X_1}} u_V\Big)\right\}\right) .$$

Since for any $\boldsymbol{\pi} \in \Omega_{\mathtt{pa}_{X_1}}$, $p_{X_1}^{\boldsymbol{\pi}}$ is a probability mass function over $X_1$, we have that $p = \sum_{X_1} p_{X_1}^{\mathtt{pa}_{X_1}} = 1$. Assume by inductive hypothesis that the result holds for $1, \ldots, i-1$, and let $\Psi_x \triangleq \bigotimes_{\Psi \in \mathcal{B}_i \setminus \mathcal{V}_0} \Psi$. Then $\Psi_i = \max([\bigotimes_{\Psi \in \mathcal{B}_i \cap \mathcal{V}_0} \Psi] \otimes \Psi_x)$. By inductive hypothesis all valuations in a set $\Psi$ in $\mathcal{B}_i \setminus \mathcal{V}_0$ have probability part $p = 1$. Hence, by definition of combination, the valuations in $\Psi_x$ contain also probability part equal to one. The reverse topological ordering implies that by the time variable $X_i$ is processed in the propagation step, all its children have been processed. Hence, the only element of $\mathcal{B}_i \cap \mathcal{V}_0$ is the set $\Psi_{X_i}$, which equals $\{(p_{X_i}^{\mathtt{pa}_{X_i}}, 0)\}$ if $X_i$ is a chance node, $\{(p_{X_i}^{\mathtt{pa}_{X_i}}, 0) : p_{X_i}^{\mathtt{pa}_{X_i}} \in \mathcal{P}_{X_i}\}$ if $X_i$ is a decision node, and $\{(1, u_{X_i})\}$ if it is a value node. Thus, we have that $\Psi_i = \max(\Psi_{X_i} \otimes \Psi_x)$. The case when $X_i$ is a value node is immediate, since any valuation in $\Psi_i$ is the result of a combination of two valuations with probability part equal to one. If $X_i$ is not a value node then

$$\begin{aligned}
\Psi_i &= \max\left(\left\{\Big(\sum_{X_i} p_{X_i}^{\mathtt{pa}_{X_i}}, \sum_{X_i} p_{X_i}^{\mathtt{pa}_{X_i}} u_x\Big) : (p_{X_i}^{\mathtt{pa}_{X_i}}, 0) \in \Psi_{\mathtt{fa}_{X_i}}, (1, u_x) \in \Psi_x\right\}\right) \\
&= \max\left(\left\{\Big(1, \sum_{X_i} p_{X_i}^{\mathtt{pa}_{X_i}} u_x\Big) : (p_{X_i}^{\mathtt{pa}_{X_i}}, 0) \in \Psi_{X_i}, (1, u_x) \in \Psi_x\right\}\right) ,
\end{aligned}$$

since $p_{X_i}^{\boldsymbol{\pi}}$ is a probability mass function for any $\boldsymbol{\pi} \in \Omega_{\mathtt{pa}_{X_i}}$. ☐

The result states that if we assume a reverse topological elimination ordering, then LVE needs to care only about the utility part of the valuations.

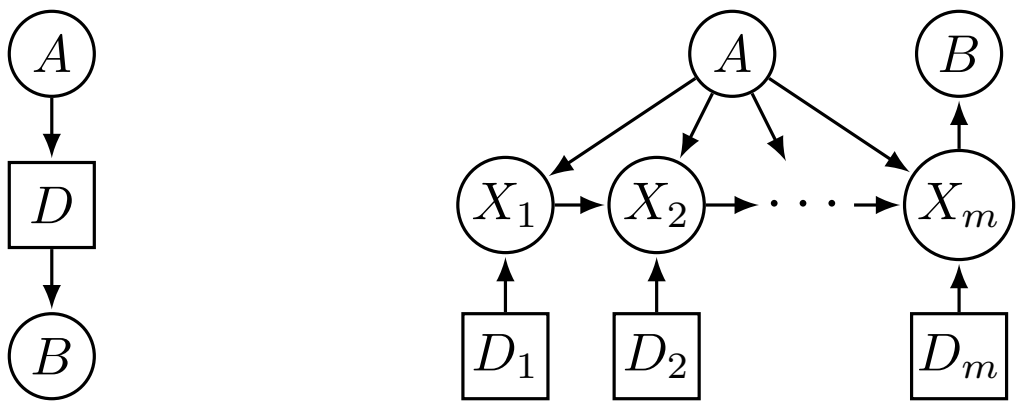

Figure 5: A simple LIMID (on the left) and its transformed MEU-equivalent (on the right).

## 3.4 Decision Nodes with Many Parents

The computational cost of adding a set $\Psi_D$ to $\mathcal{V}_0$ for a decision variable $D$ is a serious issue to the algorithm when decision nodes have many parents. To see this, consider again the example of a LIMID with a single decision node $D$ with four chance nodes as parents and one value node as child. If the decision variable has ten states, and each parent has three states, the set $\Psi_D$ contains $10^{3^4} = 10^{81}$ valuations! Fortunately, we can transform any diagram in an equivalent model in which decision nodes are parentless, and yet provides the same maximum expected utility.

**Transformation 21.** *Consider a LIMID $\mathcal{L}$. For each decision node $D$ in $\mathcal{L}$ with at least one parent, remove $D$ and add $m = |\Omega_{\mathtt{pa}_D}|$ chance nodes $X_1, \ldots, X_m$ and $m$ decision nodes $D_1, \ldots, D_m$ with domains $\Omega_{X_i} = \Omega_{D_i} = \Omega_D$ (for $i = 1, \ldots, m$). Add an arc from every parent of $D$ to each of $X_1, \ldots, X_m$, an arc from every $X_i$ to $X_{i+1}$, with $i < m$, and an arc from every $D_i$ to $X_i$ , $i = 1, \ldots, m$. Finally, add an arc from $X_m$ to each child of $D$. Assume an ordering $\boldsymbol{\pi}_1 < \cdots < \boldsymbol{\pi}_m$ of the states in $\Omega_{\mathtt{pa}_D}$. For each node $X_i$, associate a function $p_{X_i}^{\mathtt{pa}_{X_i}}$ such that for $\boldsymbol{x} \in \Omega_{\mathtt{fa}_{X_i}}$,*

$$p_{X_i}^{\mathtt{pa}_{X_i}}(\boldsymbol{x}) = \begin{cases} 1, & \textit{if } (\boldsymbol{x}^{\downarrow \mathtt{pa}_D} \neq \boldsymbol{\pi}_i \textit{ and } \boldsymbol{x}^{\downarrow X_i} = \boldsymbol{x}^{\downarrow X_{i-1}}) \\ & \textit{or } (\boldsymbol{x}^{\downarrow \mathtt{pa}_D} = \boldsymbol{\pi}_i \textit{ and } \boldsymbol{x}^{\downarrow X_i} = \boldsymbol{x}^{\downarrow X_{i-1}} = \boldsymbol{x}^{\downarrow D_i}) \\ 0, & \textit{otherwise.} \end{cases}$$

*Finally, the functions $p_X^{\mathtt{pa}_X}$ for each child $X$ of $D$ have $D$ substituted by $X_m$ in their scope, without altering the numerical values.*

Figure 5 depicts a simple LIMID with three nodes (on the left) and the diagram obtained by applying Transformation 21 (on the right). Two things are noteworthy. First, the treewidth of the transformed diagram (for any given LIMID) is increased by at most 2, because the subgraph containing the new nodes, the parents of $D$ and the children of $D$ is triangulated and contains cliques with at most $|\mathtt{pa}_D \cup \{X_i, X_{i-1}, D_i\}|$ variables.[8] Second, the functions $p_{X_i}^{\mathtt{pa}_{X_i}}$ do not correspond to sets of probability mass functions because, for instance, $\sum_{X_i} p_{X_i}^{\mathtt{pa}_{X_i}}(\boldsymbol{y}) = 0$ for $\boldsymbol{y} \in \Omega_{\mathtt{pa}_{X_i}}$ such that $\boldsymbol{y}^{\downarrow \mathtt{pa}_D} = \boldsymbol{\pi}_i$ and $\boldsymbol{y}^{\downarrow X_{i-1}} \neq \boldsymbol{y}^{\downarrow D_i}$, and hence there is $\boldsymbol{\pi} \in \Omega_{\mathtt{pa}_{X_i}}$ for which $\sum_{X_i} p^{\boldsymbol{\pi}}_{X_i}(\boldsymbol{y}) \neq 1$. Therefore, strictly speaking, the transformation does not generate a new LIMID, but, as the following result shows, a model with equal MEU that fits into the framework of LVE.

[8]Since the treewidth is given by the size of largest clique in the triangulated moral graph minus one, $|\mathtt{pa}_D|$ is a lower bound on the treewidth of the original graph.

**Proposition 22.** *Let $\mathcal{L}'$ be the result of Transformation 21 on a LIMID $\mathcal{L}$. Then*

$$\text{MEU}[\mathcal{L}'] = \text{MEU}[\mathcal{L}] .$$

The proof is in the appendix. For each decision variable $D$ in the original LIMID, the transformed model contains $m$ chance variables specifying $m|\Omega_D|^3$ values, and $m$ decision nodes with $|\Omega_D|$ states. If the treewidth of the original diagram is bounded, then $m$ is bounded by a constant and the transformation takes polynomial time. In terms of the running-time complexity of the LVE algorithm, the transformation substitutes a set $\Psi_D$ (generated in the initialization step), which would contain $|\Omega_D|^m$ valuations with scope $\mathtt{fa}_D$, with $m$ sets $\Psi_{D_i}$ containing $|\Omega_D|$ valuations with scope $D$, and $m$ singletons containing valuations with scope $\mathtt{fa}_{X_i} = \{X_i, X_{i-1}, D_i\} \cup \mathtt{pa}_D$. In the example of a ten-state decision variable with four ternary parents, the transformation generates $3^4 = 81$ sets with 10 elements each, and 81 singletons, a reduction of more than 78 orders of magnitude in the space and time required to initialize the algorithm.

If no valuation is ever discarded by the max operation in the propagation step (i.e., if all valuations generated are maximal) of LVE, after processing nodes $D_1, X_1, \ldots, X_{m-1}, D_m$ in the transformed diagram, a set containing $|\Omega_D|^m$ valuations with scope $\{X_n\} \cup \mathtt{pa}_D$ is created. This is the same number of valuations (with the same scope) that would contain the set $\Psi_D$ in $\mathcal{V}_0$ in the initialization step if LVE was run with the original diagram. Thus, the transformed diagram has a worst-case time and space complexity similar to the original diagram (with some overhead due to the increase in the number of variables). However, as the removal of non-maximal elements reduces drastically the running time and memory usage, in practice, the transformation leads to an enormous saving of computational resources.

# 4 An FPTAS For Solving Bounded LIMIDs

According to Theorems 4 and 6, solving a LIMID exactly is NP-hard even if the diagram has bounded treewidth and number of states per variable. In addition, obtaining an $\epsilon$-approximation is hard if the number of states per variable is not bounded. In this section, we show that for diagrams with bounded treewidth and number of states per variable it is possible to obtain a (multiplicative) fully polynomial time approximation scheme (FPTAS), that is, a family of $\epsilon$-approximation algorithms that runs in time polynomial in $1/\epsilon$ and in the input size.

**Definition 23.** *Given a relation $R$ on $\Psi \subseteq \Phi$, a set $\Psi'$ is called an $R$-*covering *(for $\Psi$) if for every $\phi \in \Psi$ there is $\psi \in \Psi'$ such that $\phi R \psi$.*

For example, the set $\max(\Psi)$ is a $\leq$-covering for $\Psi$. For any real number $\alpha \geq 1$, we define a relation $\leq_\alpha$ as follows.

**Definition 24.** *If $\phi = (p, u)$ and $\psi = (q, v)$ are two valuations in $\Phi$, then $\phi \leq_\alpha \psi$ if $\phi$ and $\psi$ have equal scope and $p \leq \alpha q$ and $u \leq \alpha v$.*

Notice that when $\alpha = 1$ the relation $\leq_\alpha$ as defined above is equivalent to the partial order $\leq$ in Definition 13. If $\phi$ and $\psi$ have scope $x$, deciding whether $\phi \leq_\alpha \psi$ costs at most $2|\Omega_x|$ operations of comparison of numbers. Intuitively, the relation $\leq_\alpha$ measures the maximum amount of information lost in representing $(p, u)$ by $(q, v)$, that is, $(q, v)$ approximates $(p, u)$ with a loss no greater than $\alpha$ in each of its coordinates. Notice that $\leq_\alpha$ is neither transitive nor antisymmetric, and therefore not a partial order. If $\alpha \geq \beta \geq 1$, then $\phi \leq_\beta \psi$ implies $\phi \leq_\alpha \psi$. In particular, we have that $\phi \leq \psi$ implies $\phi \leq_\alpha \psi$. Hence, any $\leq$-covering is also a $\leq_\alpha$-covering (but not the contrary).

For any real number $\alpha > 1$ we define an equivalence relation $\equiv_\alpha$ over valuations as follows.

**Definition 25.** *For any two real-valued functions $f$ and $g$ over domain $\Omega_x$, $f \equiv_\alpha g$ if for all $\boldsymbol{x} \in \Omega_x$ either $f(\boldsymbol{x}) = g(\boldsymbol{x})$ or*

$$f(\boldsymbol{x}) > 0, g(\boldsymbol{x}) > 0 \text{ and } \lfloor \log_\alpha f(\boldsymbol{x}) \rfloor = \lfloor \log_\alpha g(\boldsymbol{x}) \rfloor .$$

*If $\phi = (p, u)$ and $\psi = (q, v)$ are any two valuations, then $\phi \equiv_\alpha \psi$ if $\phi$ and $\psi$ have equal scope, and $p \equiv_\alpha q$ and $u \equiv_\alpha v$. In this case, we say that $\phi$ and $\psi$ are $\alpha$-equivalent.*

If $\phi \equiv_\alpha \psi$ then $\phi \leq_\alpha \psi$ and $\psi \leq_\alpha \phi$. Hence, for any two $\alpha$-equivalent valuations $\phi$ and $\psi$, approximating $\phi$ by $\psi$ incurs the same worst-case error as approximating $\psi$ by $\phi$. Given any finite set of valuations $\Psi$, an $\leq_\alpha$-covering for $\Psi$ can be obtained by recursively discarding any of two $\alpha$-equivalent valuations until no two $\alpha$-equivalent valuations remain in the set. We denote by $\mathrm{G}_\alpha$ an operation that returns an $\leq_\alpha$-covering for $\Psi$ in this way:

**Definition 26.** *For any finite set of valuations $\Psi \subseteq \Phi$, the operation $\mathrm{G}_\alpha$ returns a set $\mathrm{G}_\alpha(\Psi) \subseteq \Psi$ obtained by the following procedure:*

1. *Set $\mathrm{G}_\alpha(\Psi)$ initially to the empty set.*
2. *Remove an element $\phi$ from $\Psi$.*
3. *If there is* no *$\psi \in \mathrm{G}_\alpha(\Psi)$ such that $\phi \equiv_\alpha \psi$ then add $\phi$ to $\mathrm{G}_\alpha(\Psi)$. Else discard $\phi$.*
4. *If $\Psi$ is not empty, go back to step 2.*

Notice that step 2 in the definition of $\mathrm{G}_\alpha$ does not specify how an element $\phi$ from $\Psi$ should be selected. In our implementation, we randomly select a valuation to test. The following result justifies our interest in $\mathrm{G}_\alpha$.

**Lemma 27.** *If $\Psi_x$ is a finite set of valuations with scope $x$ and probability and utility part not greater than one, then $\mathrm{G}_\alpha(\Psi_x)$ is an $\leq_\alpha$-covering for $\Psi_x$ with at most $(1 - \lfloor \log_\alpha t \rfloor)^{2|\Omega_x|}$ elements, where $t$ is the smallest (strictly) positive number in the probability or utility part of a valuation in $\Psi_x$.*

*Proof.* To see that $\mathrm{G}_\alpha(\Psi_x)$ is an $\leq_\alpha$-covering for $\Psi_x$, note that, by definition of $\mathrm{G}_\alpha$, for any $(p, u) \in \Psi_x$ there is $(q, v) \in \mathrm{G}_\alpha(\Psi_x)$ such that $p \equiv_\alpha q$ and $u \equiv_\alpha v$. Hence, for all $\boldsymbol{x} \in \Omega_x$, either $p(\boldsymbol{x}) = q(\boldsymbol{x})$ or there is a negative integer $k$ such that $\alpha^k \leq p(\boldsymbol{x}) \leq \alpha^{k+1}$ and $\alpha^k \leq q(\boldsymbol{x}) \leq \alpha^{k+1}$. From this, it follows that $p \le \alpha q$. Analogously, we have that either $u(\boldsymbol{x}) = v(\boldsymbol{x})$ or $u(\boldsymbol{x}) \le \alpha^{k+1} \le \alpha v(\boldsymbol{x})$. Hence, $(p, u) \le_\alpha (q, v)$.

Let us now show that the upper bound on the cardinality of $\mathrm{G}_\alpha(\Psi_x)$ holds. For any $\boldsymbol{x} \in \Omega_x$, the $\alpha$-equivalence relation partitions the two-dimensional space $\{(p(\boldsymbol{x}), u(\boldsymbol{x})) : (p, u) \in \Psi_x\}$ in $(1 - \lfloor \log_\alpha t \rfloor)^2$ subsets of $\alpha$-equivalent valuations with respect to $\boldsymbol{x}$ (to see that, note that $p(\boldsymbol{x}) < t$ implies $p(\boldsymbol{x}) = 0$, $\lfloor \log_\alpha p(\boldsymbol{x}) \rfloor$ is an integer for $p(\boldsymbol{x}) \ge t$, and there are $-\lfloor \log_\alpha t \rfloor$ (distinct) integers between $t$ and one. The same applies for $u(\boldsymbol{x})$). Hence, $\equiv_\alpha$ partitions $\Psi_x$ in $(1 - \lfloor \log_\alpha t \rfloor)^{2|\Omega_x|}$ subsets of $\alpha$-equivalent valuations. Since $\mathrm{G}_\alpha$ discards any two $\alpha$-equivalent valuations from $\Psi_x$, $\mathrm{G}_\alpha(\Psi_x)$ have at most $(1 - \lfloor \log_\alpha t \rfloor)^{2|\Omega_x|}$ elements. □

For any LIMID with bounded treewidth and number of states per variable, it is possible to obtain in polynomial time an elimination ordering such that the cardinality of the domain of any valuation obtained by variable elimination is bounded [1, 13]. Thus, Lemma 27 guarantees that $\mathrm{G}_\alpha$ produces sets whose cardinality is polylogarithmic in the smallest positive value in the set (because the bounded treewidth and number of states imply $|\Omega_x|$ is a constant). By definition of combination and marginalization, any value in the probability or utility part of a valuation obtained during variable elimination is a polynomial on the input numbers, and so $\mathrm{G}_\alpha(\Psi_i)$ returns an $\le_\alpha$-covering for $\Psi_i$ whose cardinality is polynomially bounded by the smallest positive value (probability or scaled utility) in the input. The polynomial running time of the approximation algorithm we devise here mainly derives from this result.

For notational convenience, we define a new operation that combines set combination and $\mathrm{G}_\alpha$ as follows.

**Definition 28.** *If $\Psi_x \in \Phi_x$ and $\Psi_y \in \Phi_y$ are finite sets of valuations, we define their $\alpha$-combination $\Psi_x \oplus_\alpha \Psi_y$ for any $\alpha > 1$ as $\mathrm{G}_\alpha(\Psi_x \otimes \Psi_y)$.*

The following example shows that $\oplus_\alpha$ is not associative. Consider the following three sets of valuations over the empty domain

$$\begin{aligned}\Psi_1 &= \{(1, 0), (0.3, 0)\}\,,\\ \Psi_2 &= \{(0.5, 0), (0.1, 0)\}\,,\\ \Psi_3 &= \{(0.05, 0), (0.4, 0)\}\,,\end{aligned}$$

and assume $\mathrm{G}_\alpha$ selects always the valuation with minimum probability part among a set of $\alpha$-equivalent valuations (over the empty domain). If $\alpha = 10$, it follows that

$$\begin{aligned}(\Psi_1 \oplus_\alpha \Psi_2) \oplus_\alpha \Psi_3 &= \mathrm{G}_\alpha(\{(0.5, 0), (0.1, 0), (0.15, 0), (0.03, 0)\}) \oplus_\alpha \{(0.05, 0), (0.4, 0)\}\\ &= \{(0.1, 0), (0.03, 0)\} \oplus_\alpha \{(0.05, 0), (0.4, 0)\}\\ &= \mathrm{G}_\alpha(\{(0.005, 0), (0.04, 0), (0.0015, 0), (0.012, 0)\})\\ &= \{(0.04, 0), (0.0015, 0)\}\,.\end{aligned}$$

On the other hand, we have that

$$\begin{aligned}\Psi_1 \oplus_\alpha (\Psi_2 \oplus_\alpha \Psi_3) &= \{(1, 0), (0.3, 0)\} \oplus_\alpha \mathrm{G}_\alpha(\{(0.025, 0), (0.005, 0), (0.2, 0), (0.04, 0)\})\\ &= \{(1, 0), (0.3, 0)\} \oplus_\alpha \{(0.025, 0), (0.005, 0), (0.2, 0)\}\\ &= \{(0.025, 0), (0.0075, 0), (0.2, 0)\}\,.\end{aligned}$$

Note that associativity fails also if we identify $\alpha$-equivalent valuations:

$$\begin{aligned}\lfloor \log_\alpha([\Psi_1 \oplus_\alpha \Psi_2] \oplus_\alpha \Psi_3)\rfloor &= \{(-2,0),(-3,0)\}\\ &\neq \{(-2,0),(-3,0),(-1,0)\}\\ &= \lfloor \log_\alpha(\Psi_1 \oplus_\alpha [\Psi_2 \oplus_\alpha \Psi_3])\rfloor ,\end{aligned}$$

where $\lfloor \log_\alpha \rfloor$ is applied element-wise. Therefore, $(\Psi_1 \oplus_\alpha \Psi_2) \oplus_\alpha \Psi_3 \neq \Psi_1 \oplus_\alpha (\Psi_2 \oplus_\alpha \Psi_3)$, but, as we show in the following lemma, they are both $\leq_{\alpha^2}$-coverings for $\Psi_1 \otimes \Psi_2 \otimes \Psi_3$. It is in this last feature of $\alpha$-combination that we are mainly interested.

**Lemma 29.** *Let $a_1, \ldots, a_m$ denote nonnegative integers, and $\Psi_1, \Psi'_1, \ldots, \Psi_m, \Psi'_m$ denote finite sets of valuations such that for $i = 1, \ldots, m$, $\Psi'_i$ is a $\leq_{\alpha^{a_i}}$-covering for $\Psi_i$. Then $\Psi'_1 \oplus_\alpha \cdots \oplus_\alpha \Psi'_m$ (where the operations are applied in any order) is a $\leq_\beta$-covering for $\Psi_1 \otimes \cdots \otimes \Psi_m$, where $\beta = \alpha^{m-1+\sum_{i=1}^m a_i}$.*

*Proof.* We prove the result by induction on $k$. First notice that $\phi \leq_\alpha \psi$ is the same as $\phi \leq (\alpha, 0) \otimes \psi$. The basis ($k = 1$) follows immediately, as $\Psi'_1$ is an $\leq_{\alpha^{a_1}}$-covering for $\Psi_1$ and $\beta = \alpha^{a_1}$. Assume for $1 < k \leq m$ that $\Psi'_1 \oplus_\alpha \cdots \oplus_\alpha \Psi'_{k-1}$ is a $\leq_\gamma$-covering for $\Psi_1 \otimes \cdots \otimes \Psi_{k-1}$, where $\gamma = \alpha^{k-2+\sum_{i=1}^{k-1} a_i}$. Since $\Psi'_k$ is a $\leq_{\alpha^{a_k}}$-covering for $\Psi_k$, it follows from (A4) and the inductive hypothesis that for any $\phi \in \Psi_1 \otimes \cdots \otimes \Psi_k$ there is $\phi' \in \Psi'_1 \oplus_\alpha \cdots \oplus_\alpha \Psi'_{k-1} \otimes \Psi'_k$ such that $\phi \leq (\gamma', 0) \otimes \phi'$, where $\gamma' = \gamma\alpha^{a_k} = \alpha^{k-2+\sum_{i=1}^k a_i}$. But, by Lemma 27, for any $\phi' \in \Psi'_1 \oplus_\alpha \cdots \oplus_\alpha \Psi'_{k-1} \otimes \Psi'_k$ there is $\phi'' \in \Psi'_1 \oplus_\alpha \cdots \oplus_\alpha \Psi'_k$ such that $\phi' \leq (\alpha, 0) \otimes \phi''$. Thus, for any $\phi$ there is $\phi''$ such that $\phi \leq (\gamma', 0) \otimes \phi' \leq (\gamma', 0) \otimes (\alpha, 0) \otimes \phi''$. By transitivity of $\leq$, it follows that $\phi \leq (\beta, 0) \otimes \phi''$, where $\beta = \gamma'\alpha = \alpha^{k-1+\sum_{i=1}^k a_i}$. ∎

For any given approximation factor $\epsilon > 0$, LVE can be turned into an FPTAS by setting $\alpha = 1 + \epsilon/(2|\mathcal{U}|)$, and replacing combination of sets with $\alpha$-combination. The following algorithm, called the **$\epsilon$-LVE algorithm**, more formally describes the procedure. The only difference with respect to LVE is in the propagation step.

**Initialization:** Let $\mathcal{V}'_0$ be initially the empty set.

1. For each chance variable $X \in \mathcal{C}$, add a singleton $\Psi_X \triangleq \{(p_X^{\mathsf{pa}_X}, 0)\}$ to $\mathcal{V}_0$.
2. For each decision variable $X \in \mathcal{D}$, add a set of valuations $\Psi_X \triangleq \{(p_X^{\mathsf{pa}_X}, 0) : p_X^{\mathsf{pa}_X} \in \mathcal{P}_X\}$ to $\mathcal{V}_0$.
3. For each value variable $X \in \mathcal{V}$, add a singleton $\Psi_X \triangleq \{(1, u_X)\}$ to $\mathcal{V}_0$.

**Propagation:** For $i = 1, \ldots, n$ do:

1. Let $\mathcal{B}'_i = \emptyset$. Remove from $\mathcal{V}'_{i-1}$ all sets whose valuations contain $X_i$ in their scope and add them to $\mathcal{B}'_i$.
2. Compute $\Psi'_i \triangleq \max([\Phi_1 \oplus_\alpha \cdots \oplus_\alpha \Phi_{|\mathcal{B}'_i|}]^{-X_i})$, where, for $j = 1, \ldots, |\mathcal{B}'_i|$, $\Phi_j \in \mathcal{B}'_i$.
3. Set $\mathcal{V}'_i \triangleq \mathcal{V}'_{i-1} \cup \{\Psi'_i\}$.

**Termination:** Return $(p, u) \in \max(\bigotimes_{\Psi \in \mathcal{V}'_n} \Psi)$.

The precise order in which the $\alpha$-combinations are performed in the computation of a $\Psi'_i$ in $\epsilon$-LVE is irrelevant to the correctness of approximability results. Note, however, that different orders may lead to different solutions. For a given LIMID $\mathcal{L}$ and elimination ordering $X_1 < \cdots < X_n$, let $\Psi_i$ denote the set of valuations generated by LVE in the $i$th iteration of the propagation step, and $\Psi'_i$ its corresponding set generated by $\epsilon$-LVE. Let $s_1 \triangleq |\mathcal{B}_1| - 1 = |\mathcal{B}'_1| - 1$. For $i = 2, \ldots, n$, we define a variable $s_i$ recursively as $s_i \triangleq |\mathcal{B}_i| - 1 + \sum_{\Psi_j \in \mathcal{B}_i \setminus \mathcal{V}_0} s_j = |\mathcal{B}'_1| - 1 + \sum_{\Psi'_j \in \mathcal{B}'_i \setminus \mathcal{V}'_0} s_j$. Intuitively, $s_i$ denote the number of sets $\Psi_X$ from $\mathcal{V}_0$ that are required either directly or indirectly to compute $\Psi'_i$ (and also $\Psi_i$) minus one.

The following result is needed for the correctness of the approximation.

**Lemma 30.** *For $i \in 1, \ldots, n$, $\Psi'_i$ is a $\leq_\beta$-covering for $\Psi_i$, where $\beta = \alpha^{s_i}$.*

*Proof.* We prove the result by induction on $i$. Since $\mathcal{B}_1 = \mathcal{B}'_1$, it follows from Lemma 29 with $a_1 = \cdots = a_{|\mathcal{B}_1|} = 0$ that $[\bigoplus_{\Psi \in \mathcal{B}'_1} \Psi]^{-X_1}$ is a $\leq_\beta$-covering for $[\bigotimes_{\Psi \in \mathcal{B}_1} \Psi]^{-X_1}$, where $\beta = \alpha^{s_1}$. Hence, for any $\phi \in \Psi_1$ there is $\phi' \in [\bigoplus_{\Psi \in \mathcal{B}'_1} \Psi]^{-X_1}$ such that $(\alpha^{s_1}, 0) \otimes \phi' \geq \phi$. But for any $\phi'$ there is $\phi'' \in \Psi'_1$ such that $\phi'' \geq \phi'$. Hence, the basis follows from $(\alpha^{s_1}, 0) \otimes \phi'' \geq (\alpha^{s_1}, 0) \otimes \phi' \geq \phi$. Assume the result holds for $j = 1, \ldots, i-1$, and let $m = |\mathcal{B}_i| = |\mathcal{B}'_i|$. The set $\Psi_i$ equals $\max([\Phi_1 \otimes \cdots \otimes \Phi_{|\mathcal{B}_i|}]^{-X_i})$, where each $\Phi_k$ either equals some $\Psi_j$, with $j < i$, or is in $\mathcal{V}_0$. Likewise, $\Psi'_i = \max([\Phi'_1 \oplus_\alpha \cdots \oplus_\alpha \Phi'_{|\mathcal{B}_i|}]^{-X_i})$, where each $\Phi'_k$ either equals some $\Psi'_j$ (so that $\Phi_k = \Psi_j$ implies $\Phi'_k = \Psi'_j$ and vice-versa) or is in $\mathcal{V}'_0 = \mathcal{V}_0$. For $k = 1, \ldots, m$, let $a_k = s_j$ if $\Phi_k = \Psi_j$ (and $\Phi'_k = \Psi'_j$) and $a_k = 0$ if $\Phi_k \in \mathcal{V}_0$. By inductive hypothesis, each $\Phi'_k$ is a $\leq_{\alpha^{a_k}}$-covering of $\Phi_k$ (if $\Phi'_k \in \mathcal{V}'_0$ then $\Phi'_k$ equals $\Phi_k$ and it is therefore a $\leq$-covering for $\Phi_k$). Then, by Lemma 29, $\Phi'_1 \oplus_\alpha \cdots \oplus_\alpha \Phi'_m$ is a $\leq_\beta$-covering for $\Phi_1 \otimes \cdots \otimes \Phi_m$, where $\beta = \alpha^{m-1+\sum_{k=1}^m a_k} = \alpha^{s_i}$. Also, by (A5) $[\Phi'_1 \oplus_\alpha \cdots \oplus_\alpha \Phi'_{|\mathcal{B}_i|}]^{-X_i}$ is a $\leq_\beta$-covering for $[\Phi_1 \otimes \cdots \otimes \Phi_{|\mathcal{B}_i|}]^{-X_i}$. Finally, the result follows from transitivity of the partial order, that is, $\Psi'_i$ is a $\leq_\beta$-covering for $\Psi_i$. ☐

The $\alpha$-combinations make sure that the cardinality of the sets remains bounded during the propagation. Hence, if the diagram has bounded treewidth and number of states per variable, the following result guarantees the correctness of the approximation and the polynomial running time in the input length and $1/\epsilon$.

**Theorem 31.** *If $\mathcal{L}$ is a LIMID with bounded treewidth and number of states per variable, then $\epsilon$-LVE is an FPTAS for* MEU$[\mathcal{L}]$.

*Proof.* First, we show that $\epsilon$-LVE indeed obtains an $\epsilon$-approximation to MEU$[\mathcal{L}]$.

Let $\mathcal{F}_1 \triangleq \mathcal{B}_1$. For $i = 2, \ldots, n$, let $\mathcal{F}_i \triangleq (\mathcal{B}_i \cap \mathcal{V}_0) \cup \bigcup_{\Psi_j \in \mathcal{B}_i \setminus \mathcal{V}_0} \mathcal{F}_j$ denote the collection of sets $\Psi \in \mathcal{V}_0$ that were used directly or indirectly in the computation of $\Psi_i$. Then $s_1 = |\mathcal{F}_1| - 1$. Assume by induction that $s_j = |\mathcal{F}_j| - 1$ for $j = 1, \ldots, i-1$. Hence, $s_i = |\mathcal{B}_i| - 1 + \sum_{\Psi_j \in \mathcal{B}_i \setminus \mathcal{V}_0} |\mathcal{F}_j| - 1$, which equals $|\mathcal{B}_i| - 1 - |\mathcal{B} \setminus \mathcal{V}_0| + \sum_{\Psi_j \in \mathcal{B}_i \setminus \mathcal{V}_0} |\mathcal{F}_j|$. But $|\mathcal{B}_i| - |\mathcal{B}_i \setminus \mathcal{V}_0|$ equals $|\mathcal{B}_i \cap \mathcal{V}_0|$. Thus, $s_i = |\mathcal{B}_i \cap \mathcal{V}_0| - 1 + \sum_{\Psi_j \in \mathcal{B}_i \setminus \mathcal{V}_0} |\mathcal{F}_j|$, which equals $|\mathcal{F}_i| - 1$ because the sets in $\mathcal{V}_0$ are used exactly once (hence the sets $\mathcal{F}_j$ for different $\Psi_j \in \mathcal{B}_i \setminus \mathcal{V}_0$ are disjoint, and also $\mathcal{B}_i \cap \mathcal{V}_0$).

Let $m = |\mathcal{U}|$. The collections $\mathcal{V}_n$ and $\mathcal{V}'_n$ contain only sets $\Psi_i$ and $\Psi'_i$, respectively, generated during the propagation. Thus, $\sum_{\Psi_i \in \mathcal{V}'_n} s_i = m - |\mathcal{V}_n|$, because there are $|\mathcal{U}|$ sets in $\mathcal{V}_0$ and each set belongs to exactly one $\mathcal{F}_i$ for $\Psi_i \in \mathcal{V}_n$. Like in the exact case (i.e., in LVE), the valuations in $\bigotimes_{\Psi'_i \in \mathcal{V}'_n} \Psi'_i$ are pairs $(1, \mathrm{E}_s[\mathcal{L}])$ for some strategy $s$. Hence, $\max(\bigotimes_{\Psi_i \in \mathcal{V}_n} \Psi_i)$ and $\max(\bigotimes_{\Psi'_i \in \mathcal{V}'_n} \Psi'_i)$ return each a single valuation. Consider the valuation $\phi^* \in \max(\bigotimes_{\Psi_i \in \mathcal{V}_n} \Psi_i)$. By definition, $\phi^*$ factorizes as $\bigotimes_i \phi^*_i$, where each $\phi^*_i$ belongs to exactly one set $\Psi_i$ in $\mathcal{V}_n$. According to Lemma 30, for each $\phi^*_i$ there is $\phi_i \in \Psi'_i$ such that $\phi^*_i \leq (\alpha^{s_i}, 0) \otimes \phi_i$. Thus, it follows from (A4) that

$$\phi^* = \bigotimes_{\Psi_i \in \mathcal{V}_n} \phi^*_i \leq \bigotimes_{\Psi'_i \in \mathcal{V}'_n} [(\alpha^{s_i}, 0) \otimes \phi_i] \, .$$

By associativity of $\otimes$, we have that

$$\begin{aligned} \bigotimes_{\Psi'_i \in \mathcal{V}'_n} [(\alpha^{s_i}, 0) \otimes \phi_i] &= (\alpha^{\sum s_i}, 0) \otimes \left( \bigotimes_{\Psi'_i \in \mathcal{V}'_n} \phi_i \right) \\ &= (\alpha^{m - |\mathcal{V}_n|}, 0) \otimes \left( \bigotimes_{\Psi'_i \in \mathcal{V}'_n} \phi_i \right) . \end{aligned}$$

Let $\phi = \bigotimes_{\Psi_i \in \mathcal{V}'_n} \phi_i$. Hence, $\phi^* \leq_{\alpha^{m-|\mathcal{V}_n|}} \phi$, which implies $\phi^* \leq_{\alpha^m} \phi$ and therefore $\alpha^m u \geq u^*$, where $u$ and $u^*$ denote the utility part of $\phi$ and $\phi^*$, respectively. If $\phi$ is not in $\max(\bigotimes_{\Psi_i \in \mathcal{V}'_n} \Psi'_i)$, then there is $\phi' \in \max(\bigotimes_{\Psi_i \in \mathcal{V}'_n} \Psi'_i)$ such that $\phi \leq \phi'$, and thus $\phi^* \leq_{\alpha^m} \phi'$. Thus, we can assume without loss of generality that $\phi \in \max(\bigotimes_{\Psi_i \in \mathcal{V}'_n} \Psi'_i)$, so that $\epsilon$-LVE outputs the utility part $u$ of $\phi$. Since $\alpha = 1 + \epsilon/(2m)$, it follows from Lemma 37 (in the appendix) that

$$(1 + \epsilon) u \geq (1 + \epsilon/2m)^m u \geq u^* \, .$$

Let us now analyze the time complexity of the algorithm. Let $\omega - 1$ denote the tree-width of the network and $\kappa$ the maximum number of states of a variable, both considered bounded. Hence, for any variable in $\mathcal{U}$ we have that $|\mathtt{fa}_X| \leq \omega$ and $|\Omega_{\mathtt{fa}_X}| \leq \kappa^\omega$. The initialization step takes then $O(m)$ time:

$$O\left( \sum_{X \in \mathcal{C} \cup \mathcal{V}} 2|\Omega_{\mathtt{fa}_X}| \right) \leq O(|\mathcal{C} \cup \mathcal{V}| \kappa^\omega) \leq O(|\mathcal{C} \cup \mathcal{V}|) \leq O(m)$$

time to generate the sets associated to chance and value nodes, and

$$O\left( \sum_{D \in \mathcal{D}} 2|\Omega_{\mathtt{fa}_D}||\Omega_D|^{|\Omega_{\mathtt{pa}_D}|} \right) \leq O(|\mathcal{D}| \kappa^\omega \kappa^{\kappa^\omega}) \leq O(|\mathcal{D}|) \leq O(m)$$

time to generate the sets associated to decision nodes.

Let us consider the propagation step. First, we need to find all sets containing a variable $X_i$. Since each set has a scope with at most $\omega$ variables, and there are $O(m)$ sets in $\mathcal{V}_{i-1}$ (the $i-1$ sets $\Psi_j$ generated in the previous iterations plus the $O(|\mathtt{ch}_{X_i} \cup \{X_i\}|) \leq O(m)$ sets from $\mathcal{V}_0 \setminus \bigcup_{j=1}^{i-1} \mathcal{V}_j$), the set $\mathcal{B}'_i$ can be obtained in $O(m\omega) \leq O(m)$ time.

To compute $\Psi'_i$, we first have to compute the set of marginals $(\bigoplus_{\Psi \in \mathcal{B}'_i} \Psi)^{-X_i}$ and then obtain the set of maximal valuations. To compute the former we need to perform $|\mathcal{B}'_i| - 1$ operations $\mathrm{G}_\alpha(\Phi_1 \otimes \Phi_2)$ and a marginalization, where $\Phi_1$ and $\Phi_2$ are sets either in $\mathcal{V}'_0$ or equal to some $\Psi_j$, with $j < i$. Each set $\Psi$ in $\mathcal{B}'_i \cap \mathcal{V}'_0$ contains $O(\kappa^\omega) \leq O(1)$ elements.

We will obtain a bound for the number of elements in some

$$\Psi'_j = \max([\mathrm{G}_\alpha([\Phi])]^{-X_j}) \,,$$

where $\Phi = \bigoplus_{\Psi \in \mathcal{B}'_j} \Psi$. Let $t$ denote the smallest positive number in the probability or in the utility part of a valuation in $\Psi'_j$, and $b$ denote the number of bits required to encode $\mathcal{L}$. Since the input probabilities and utilities are rational numbers, each positive input number is not smaller than $2^{-b}$ (otherwise we would need more than $b$ bits to encode it). The valuations in $\Psi'_j$ can be obtained by a sequence of marginalizations and combinations of valuations, where each valuation is in some $\Psi \in \mathcal{F}_j \subseteq \mathcal{V}_0$. Hence, $t$ is obtained by a series of multiplications and additions and it is therefore a polynomial on the input numbers (the probabilities and utilities associated to chance and value variables in $\mathcal{L}$). For each variable $X \in \mathcal{C} \cup \mathcal{V}$ there are $O(\kappa^\omega)$ input numbers. Therefore $t$ is a polynomial of degree $O(m\kappa^\omega) \leq O(m)$. Since the inputs of the polynomial are either zero or some number greater than or equal to $2^{-b}$, it follows that $t \geq 2^{-bO(m)}$. Let $x$ denote the scope of $\Psi'_j$. Since $\Psi'_j = \max([\mathrm{G}_\alpha([\Phi)]^{-X_j})$, it follows from Lemma 27 that $\Psi$ contains $O((1 - \lfloor \log_\alpha t \rfloor)^{2|\Omega_x|}) \leq O([bm/\ln(\alpha)]^{2\omega})$ elements. Since $\alpha = 1 + \epsilon/2m$, we have from Lemma 38 (in the appendix) that

$$O\left(\left[\frac{bm}{\ln(\alpha)}\right]^{2\omega}\right) \leq O\left(\left[bm\frac{1+\epsilon/2m}{\epsilon/2m}\right]^{2\omega}\right) .$$

Hence, the number of valuations in any $\Psi'_j$ is $O([bm^2/\epsilon]^{2\omega})$, which is polynomial in $b$, $m$ and $1/\epsilon$.

The set $(\bigoplus_{\Psi \in \mathcal{B}'_i} \Psi)^{-X_i}$ can thus be obtained by

$$O\left((|\mathcal{B}'_i| - 1)\left(\left[\frac{bm^2}{\epsilon}\right]^{2\omega}\right)^2\right) \leq O\left(\left[\frac{b}{\epsilon}\right]^{2\omega} m^{2\omega+2}\right)$$

combinations and $O([bm^2/\epsilon]^{2\omega})$ marginalizations. The set of maximal valuations can be obtained by pairwise comparison of all valuations, and thus also takes polynomial time. Finally, computing a set $\mathcal{V}'_i$ takes time proportional to the number of sets in $\mathcal{V}'_{i-1}$ and $\mathcal{B}'_i$, which is a polynomial on $m$. □

# 5 Experiments

We evaluate the performance of the algorithms on random LIMIDs generated in the following way. Each LIMID is parameterized by the number of decision nodes $d$, the number of chance nodes $c$, the maximum cardinality of the domain of a chance variable family $\omega_C$, and the maximum cardinality of the domain of a decision variable family $\omega_D$. We set the number of value nodes $v$ to be $d + 2$. For each variable $X_i$, $i = 1, \ldots, c + d + v$, we sample $\Omega_{X_i}$ to contain from 2 to 4 states. Then we repeatedly add an arc from a decision node with no children to a value node with no parents (so that each decision node has at least one value node as children). This step guarantees that all decisions are relevant for the computation of the MEU. Finally, we repeatedly add an arc that neither makes the domain of a variable greater than the given bounds nor makes the treewidth more than 10, until no arcs can be added without exceeding the bounds.[9] Note that this generates diagrams where decision and chance variables have at most $\log_2 \omega_D - 1$ and $\log_2 \omega_C - 1$ parents, respectively. Once the DAG is obtained, we randomly sample the probability mass functions and utility functions associated to chance and value variables, respectively.

We compare LVE against the CR algorithm of de Campos and Ji [5] in 1620 LIMIDs randomly generated by the described procedure with parameters $5 \leq d \leq 50$, $8 \leq c \leq 50$, $8 \leq \omega_D \leq 64$ and $16 \leq \omega_C \leq 64$. LVE was implemented in C++ and tested in the same computer as CR.[10] Table 1 contrasts the running times of each algorithm (averages ± standard deviation) for different configurations of randomly generated LIMIDs. Each row contains the percentage of solved diagrams ($S_{\text{CR}}$ and $S_{\text{LVE}}$) and time performance ($T_{\text{CR}}$ and $T_{\text{LVE}}$) of each of the algorithms for $N$ diagrams randomly generated using parameters $d$, $c$, $v$, $\omega_D$, and $\omega_C$. For each fixed parameter configuration, LVE outperforms CR by orders of magnitude. Also, CR was unable to solve most of the diagrams with more than 50 variables, whereas LVE could solve diagrams containing up to 150 variables and with $\omega_D \leq 32$. Both algorithms failed to solve diagrams $\omega_D = 64$. A diagram is consider unsolved by an algorithm if the algorithm was not able to reach the exact solution within the limit of 12 hours. All in all, LVE appears to scale well on the number of nodes (i.e., on $d$, $c$ and $v$) but poorly on the domain cardinality of the family of decision variables (i.e., on $\omega_D$).

A good succinct measure of the hardness of solving a LIMID is the total number of strategies $|\Delta|$, which represents the size of the search space in a brute-force approach. $|\Delta|$ can also be loosely interpreted as the total number of alternatives (over all decision variables) in the problem instance. Figure 6 depicts running time against number of strategies in a log-log scale for the two algorithms on the same test set of random diagrams. For each algorithm, only solved instances are shown, which covers approximately 96% of the cases for LVE, and 68% for CR. We note that LVE solved all cases that CR solved (but not the opposite). Again, we see that LVE is orders of magnitude faster than CR. Within the limit of 12 hours, LVE was able to compute diagrams containing up to $10^{64}$ strategies, whereas CR solved diagrams with at most $10^{25}$ strategies.

The reduction in complexity obtained by the removal of non-maximal valuations during the propagation step can be checked in Figure 7, which shows the maximum cardinality of a set $\Psi_i$ generated in the propagation step in contrast to the number of strategies. For each diagram (a point in the figure) solved by LVE the cardinality of the sets remains bounded above by $10^6$ while we vary the number of strategies (which equals the largest cardinality of a propagated set in the worst case where no valuation is discarded). This shows that the worst-case

[9] Since current algorithms for checking whether the treewidth of a graph exceeds a fixed $k$ are too slow for $k \geq 5$ [1], we resort to a greedy heuristic that resulted in diagrams whose actual treewidth ranged from 5 to 10.

[10] We used the CR implementation available at `http://www.idsia.ch/~cassio/id2mip/` and CPLEX [17] as mixed integer programming solver. Our LVE implementation and the test cases are available at `http://www.idsia.ch/~cassio/lve/`.

| *id* | $N$ | $d$ | $c$ | $v$ | $\omega_D$ | $\omega_C$ | $S_{\text{CR}}$ (%) | $T_{\text{CR}}$ (s) | $S_{\text{LVE}}$ (%) | $T_{\text{LVE}}$ (s) |
|---|---|---|---|---|---|---|---|---|---|---|
| *1* | 60 | 5 | 8 | 7 | 12 | 16 | 100 | $6 \pm 45$ | 100 | $0.006 \pm 0.01$ |
| *2* | 60 | 5 | 8 | 7 | 16 | 16 | 100 | $9 \pm 43$ | 100 | $0.02 \pm 0.05$ |
| *3* | 60 | 5 | 8 | 7 | 8 | 16 | 100 | $6 \pm 51$ | 100 | $0.002 \pm 0.01$ |
| *4* | 60 | 10 | 8 | 12 | 12 | 16 | 98 | $15 \pm 53$ | 100 | $0.02 \pm 0.02$ |
| *5* | 60 | 10 | 8 | 12 | 16 | 16 | 93 | $107 \pm 273$ | 100 | $103 \pm 786$ |
| *6* | 60 | 10 | 8 | 12 | 8 | 16 | 100 | $0.4 \pm 0.2$ | 100 | $0.007 \pm 0.01$ |
| *7* | 60 | 10 | 28 | 12 | 12 | 16 | 96 | $1175 \pm 6126$ | 100 | $0.05 \pm 0.08$ |
| *8* | 60 | 10 | 28 | 12 | 16 | 16 | 83 | $3340 \pm 8966$ | 100 | $0.2 \pm 0.2$ |
| *9* | 30 | 10 | 28 | 12 | 16 | 64 | 10 | $2838 \pm 1493$ | 96 | $47 \pm 142$ |
| *10* | 30 | 10 | 28 | 12 | 32 | 16 | 93 | $1070 \pm 2461$ | 100 | $0.2 \pm 0.4$ |
| *11* | 60 | 10 | 28 | 12 | 32 | 32 | 0 | — | 93 | $905 \pm 2847$ |
| *12* | 30 | 10 | 28 | 12 | 32 | 64 | 3 | $73 \pm 0$ | 86 | $2440 \pm 7606$ |
| *13* | 30 | 10 | 28 | 12 | 64 | 64 | 0 | — | 0 | — |
| *14* | 60 | 10 | 28 | 12 | 8 | 16 | 100 | $1 \pm 3$ | 100 | $0.01 \pm 0.007$ |
| *15* | 60 | 20 | 8 | 22 | 12 | 16 | 93 | $2687 \pm 7564$ | 100 | $155 \pm 1196$ |
| *16* | 90 | 20 | 8 | 22 | 16 | 16 | 38 | $5443 \pm 10070$ | 98 | $270 \pm 1822$ |
| *17* | 30 | 20 | 8 | 22 | 16 | 64 | 30 | $9660 \pm 10303$ | 100 | $29 \pm 84$ |
| *18* | 60 | 20 | 8 | 22 | 32 | 32 | 0 | — | 78 | $938 \pm 1417$ |
| *19* | 30 | 20 | 8 | 22 | 32 | 64 | 0 | — | 76 | $1592 \pm 3402$ |
| *20* | 30 | 20 | 8 | 22 | 64 | 64 | 0 | — | 0 | — |
| *21* | 60 | 20 | 8 | 22 | 8 | 16 | 100 | $7 \pm 20$ | 100 | $0.02 \pm 0.008$ |
| *22* | 60 | 10 | 78 | 12 | 16 | 16 | 60 | $5944 \pm 9920$ | 100 | $0.5 \pm 0.5$ |
| *23* | 30 | 10 | 78 | 12 | 32 | 16 | 70 | $3820 \pm 8127$ | 100 | $0.6 \pm 1$ |
| *24* | 60 | 20 | 58 | 22 | 12 | 16 | 50 | $6455 \pm 9344$ | 100 | $522 \pm 4011$ |
| *25* | 60 | 20 | 58 | 22 | 16 | 16 | 11 | $11895 \pm 12662$ | 100 | $2 \pm 11$ |
| *26* | 60 | 20 | 58 | 22 | 8 | 16 | 96 | $849 \pm 4098$ | 100 | $0.07 \pm 0.04$ |
| *27* | 60 | 30 | 38 | 32 | 12 | 16 | 28 | $3416 \pm 4827$ | 98 | $35 \pm 214$ |
| *28* | 30 | 30 | 38 | 32 | 16 | 16 | 0 | — | 100 | $2 \pm 10$ |
| *29* | 60 | 30 | 38 | 32 | 8 | 16 | 96 | $2261 \pm 6572$ | 100 | $0.1 \pm 0.03$ |
| *30* | 30 | 30 | 88 | 32 | 12 | 16 | 0 | — | 100 | $230 \pm 1027$ |
| *31* | 30 | 30 | 88 | 32 | 8 | 16 | 60 | $3448 \pm 5837$ | 100 | $0.2 \pm 0.1$ |
| *32* | 30 | 50 | 48 | 52 | 12 | 16 | 0 | — | 96 | $1753 \pm 7405$ |
| *33* | 30 | 50 | 48 | 52 | 8 | 16 | 10 | $5014 \pm 2974$ | 100 | $0.5 \pm 0.09$ |

Table 1: Performance of LVE and CR on randomly generated LIMIDs (numbers are rounded down).

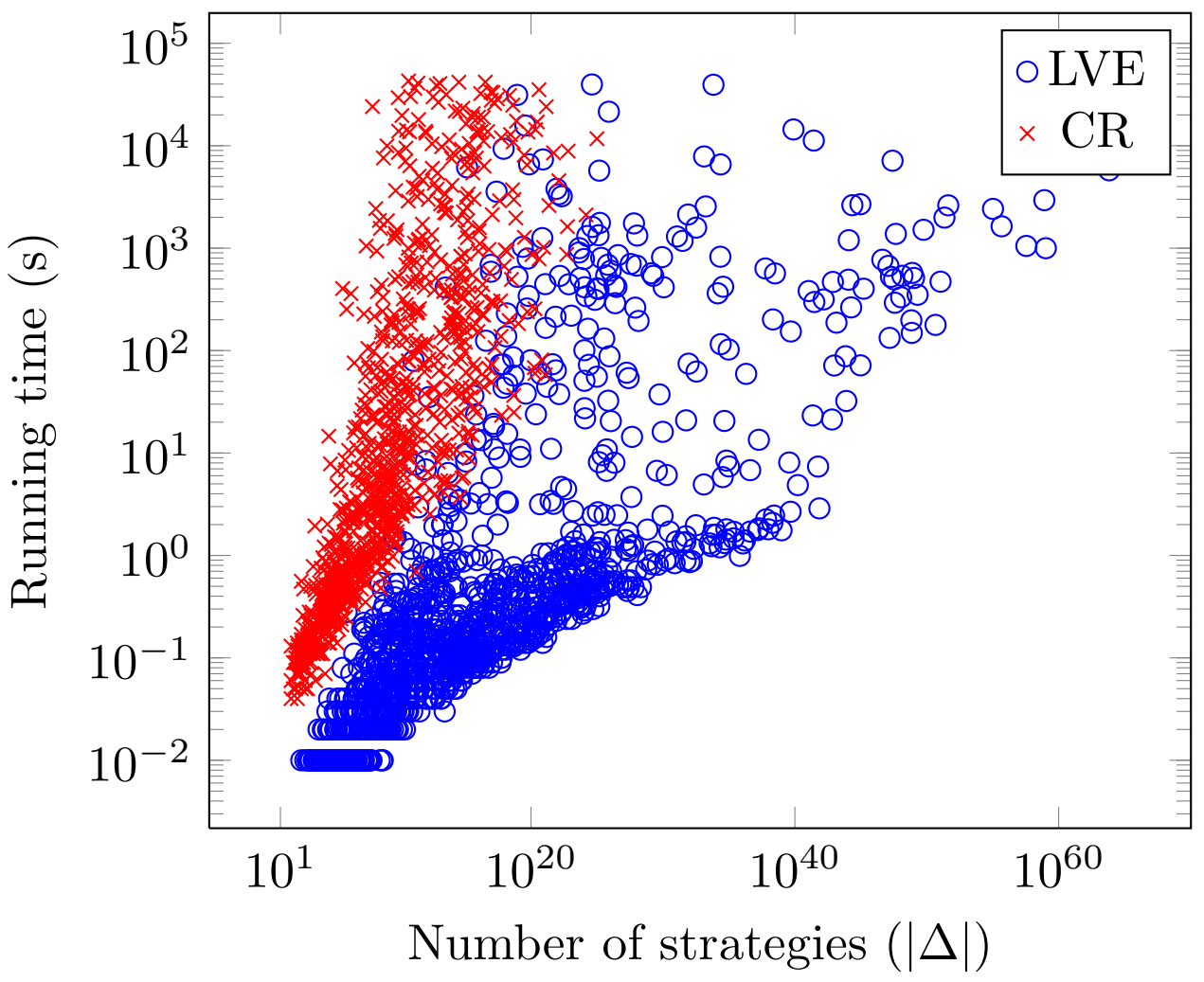

Figure 6: Running time of LVE and CR on randomly generated LIMIDs.

analysis in Section 3.1 is excessively pessimistic.

We also compared the performance of LVE with its FPTAS version $\epsilon$-LVE. The results using approximation factors $\epsilon = 0.1$ and $\epsilon = 0.01$ are in Table 2. The numbers in the first column identify each row by the corresponding row in Table 1. The second and the fifth columns describe the percentage $S(\epsilon)$ of instances solved by $\epsilon$-LVE within 12 hours for different approximation factors $\epsilon$; third and sixth columns report the average and standard deviation of relative running time on instances which both $\epsilon$-LVE and LVE were able to solve within the time limit, that is,

$$\Delta T(\epsilon) \triangleq \frac{1}{n(\epsilon)} \sum \frac{T(\epsilon) - T_{\text{LVE}}}{T_{\text{LVE}}} \pm \sqrt{\frac{1}{n(\epsilon)} \sum \left( \frac{T(\epsilon) - T_{\text{LVE}}}{T_{\text{LVE}}} \right)^2 - \left( \frac{1}{n(\epsilon)} \sum \frac{T(\epsilon) - T_{\text{LVE}}}{T_{\text{LVE}}} \right)^2},$$

where $T(\epsilon)$ denotes the running time of $\epsilon$-LVE (run with approximation factor $\epsilon$) on an instance, $n(\epsilon)$ denotes the number of cases solved by both $\epsilon$-LVE and LVE, and the sums are over these cases. Negative values of $\Delta T$ denote (sets of) instances on which $\epsilon$-LVE was faster than LVE. Finally, the fourth and the last columns show the relative maximum cardinality of a set in $\epsilon$-LVE with respect to LVE:

$$\Delta C(\epsilon) \triangleq \frac{1}{n(\epsilon)} \sum \frac{C(\epsilon) - C_{\text{LVE}}}{C_{\text{LVE}}} \pm \sqrt{\frac{1}{n(\epsilon)} \sum \left( \frac{C(\epsilon) - C_{\text{LVE}}}{C_{\text{LVE}}} \right)^2 - \left( \frac{1}{n(\epsilon)} \sum \frac{C(\epsilon) - C_{\text{LVE}}}{C_{\text{LVE}}} \right)^2},$$

where $C(\epsilon)$ denote the maximum cardinality of a set $\Psi'_i$ produced by $\epsilon$-LVE with approximation factor $\epsilon$, and $C_{\text{LVE}} = \max_i |\Psi_i|$. As with $\Delta T$, negative values of $\Delta C$ indicate cases in which $\epsilon$-LVE produced (on average) sets smaller in cardinality than those produced by LVE. From the table, we see that the approximation algorithm is slower (on average) than the exact version apart from three sets of instances (viz. sets 6, 30, 31, 32) with approximation factor 0.1. We credit the inferior performance of $\epsilon$-LVE to the extra complexity introduced by the $G_\alpha$ operations. Additionally, as Theorem 6 shows, the polynomial running time of $\epsilon$-LVE can be obtained only by considering the cardinality of domains bounded, which in our experiments signifies a low value of $\max\{\omega_D, \omega_C\}$. To see that the $G_\alpha$ operation indeed remove elements from the sets, note that the maximum cardinality of a set $\Psi'_i$ produced by $\epsilon$-LVE is smaller (on average) than the maximum cardinality of a set $\Psi_i$ produced by LVE on almost all instances (i.e., $\Delta C$ is negative). Finally, we note that with $\epsilon = 0.1$, $\epsilon$-LVE was able to solve six cases which LVE failed to solved within the time limit, whereas LVE solved two cases which $\epsilon$-LVE could not solve. For $\epsilon = 0.01$, the exact version was able to solve 14 cases where the approximation failed, and $\epsilon$-LVE solved five cases which LVE was not able to solve. All cases solved by $\epsilon$-LVE with $\epsilon = 0.01$ were also solved with $\epsilon = 0.1$ (but not the converse).

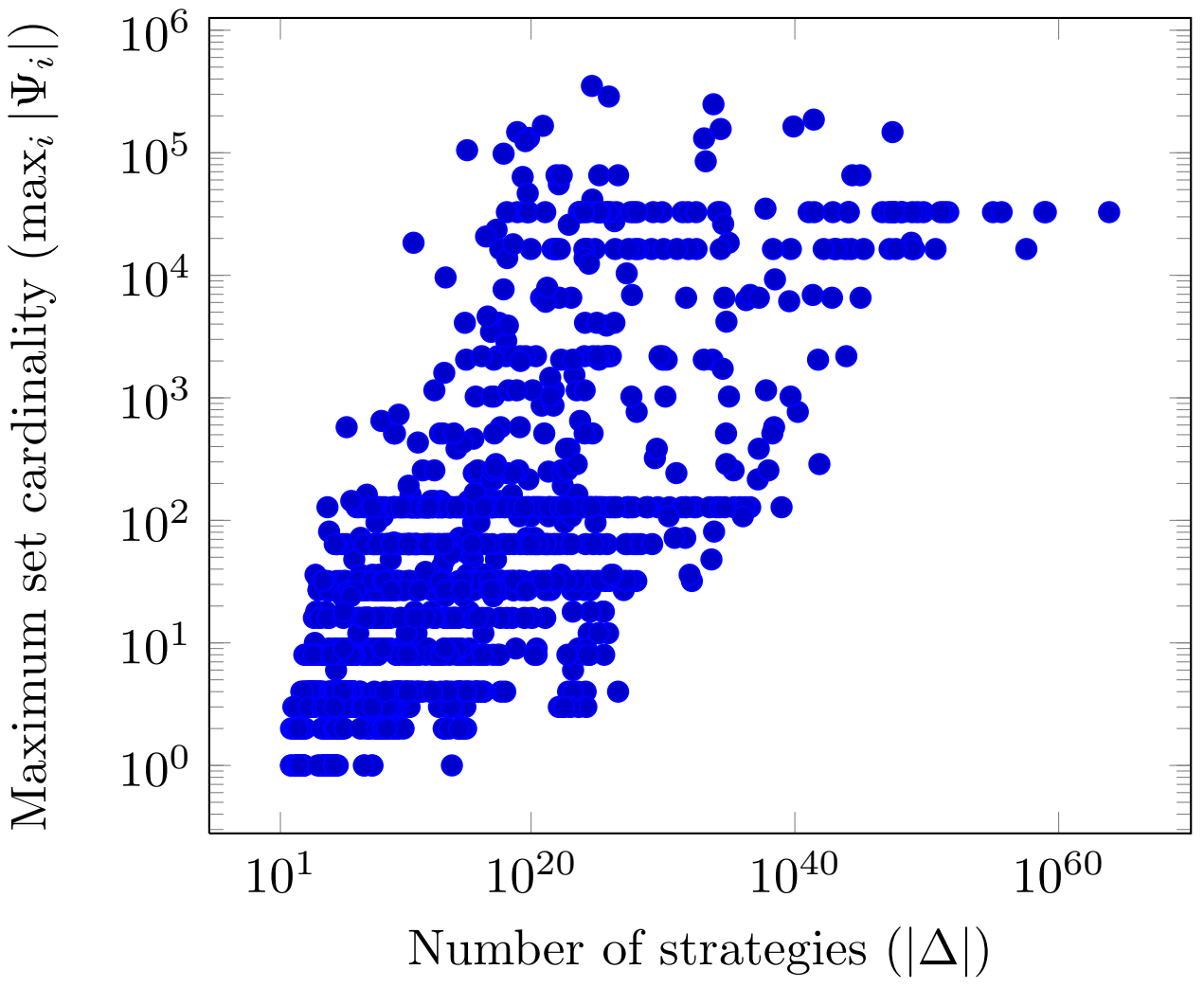

Figure 7: Maximum number of valuations in a set during the propagation step of LVE.

# 6 Related Work

Influence diagrams were introduced by Howard and Matheson [12] as a concise language for the specification of utility-based decision problems. There is a substantial literature that formalizes influence diagrams and develop algorithms under the premises of no forgetting and regularity [2, 20, 21]. We point the interested reader to the works of Jensen and Nielsen [13] and Koller and Friedman [15].

Zhang et al. [23] studied families of LIMIDs that could be solved by dynamic programming, such as LIMIDS respecting no forgetting and regularity. The SPU algorithm of Lauritzen and Nilsson [16] solves these cases in polynomial time if the diagram has bounded treewidth. To the best of our knowledge, the only attempt to (globally) solve arbitrary LIMIDs exactly without recurring to an exhaustive search on the space of strategies is the CR algorithm of de Campos and Ji [5] against which we compare our algorithm.

| *id* | $S(\epsilon = 0.1)$ | $\Delta T(\epsilon = 0.1)$ | $\Delta C(\epsilon = 0.1)$ | $S(\epsilon = 0.01)$ | $\Delta T(\epsilon = 0.01)$ | $\Delta C(\epsilon = 0.01)$ |
|---|---|---|---|---|---|---|
| *1* | 100 | 0.13 ± 0.43 | -0.15 ± 0.28 | 100 | 1.48 ± 2.96 | -0.15 ± 0.28 |
| *2* | 100 | 0.27 ± 0.57 | -0.11 ± 0.24 | 100 | 3.16 ± 4.70 | -0.11 ± 0.24 |
| *3* | 100 | 0.01 ± 0.11 | -0.12 ± 0.24 | 100 | 0.39 ± 1.41 | -0.12 ± 0.24 |
| *4* | 100 | 0.24 ± 0.49 | -0.18 ± 0.31 | 100 | 2.58 ± 3.16 | -0.18 ± 0.31 |
| *5* | 100 | 0.59 ± 0.83 | -0.13 ± 0.31 | 100 | 6.60 ± 6.47 | -0.13 ± 0.31 |
| *6* | 100 | -0.02 ± 0.10 | -0.18 ± 0.28 | 100 | 0.18 ± 0.49 | -0.18 ± 0.28 |
| *7* | 100 | 0.41 ± 0.56 | -0.12 ± 0.25 | 100 | 4.25 ± 4.55 | -0.12 ± 0.25 |
| *8* | 100 | 1.22 ± 0.90 | 0.00 ± 0.00 | 100 | 11.92 ± 9.01 | 0.00 ± 0.00 |
| *9* | 96 | 1.94 ± 1.04 | 0.00 ± 0.00 | 96 | 19.94 ± 10.64 | 0.00 ± 0.00 |
| *10* | 100 | 1.08 ± 1.10 | 0.00 ± 0.00 | 100 | 10.64 ± 10.57 | 0.00 ± 0.00 |
| *11* | 93 | 1.55 ± 1.13 | -0.02 ± 0.12 | 90 | 15.71 ± 9.78 | -0.02 ± 0.12 |
| *12* | 80 | 1.54 ± 0.91 | -0.01 ± 0.05 | 73 | 15.90 ± 9.16 | -0.01 ± 0.05 |
| *13* | 0 | — | — | 0 | — | — |
| *14* | 100 | 0.07 ± 0.34 | -0.14 ± 0.24 | 100 | 0.81 ± 1.62 | -0.14 ± 0.24 |
| *15* | 100 | 0.18 ± 0.35 | -0.23 ± 0.35 | 100 | 2.58 ± 2.30 | -0.23 ± 0.35 |
| *16* | 98 | 0.91 ± 1.19 | -0.24 ± 0.40 | 98 | 10.37 ± 9.86 | -0.24 ± 0.40 |
| *17* | 100 | 1.07 ± 1.05 | -0.14 ± 0.30 | 100 | 11.64 ± 9.69 | -0.14 ± 0.30 |
| *18* | 78 | 1.69 ± 1.36 | -0.02 ± 0.12 | 66 | 15.99 ± 13.47 | -0.03 ± 0.13 |
| *19* | 76 | 1.45 ± 1.65 | -0.16 ± 0.31 | 73 | 15.07 ± 14.62 | -0.17 ± 0.31 |
| *20* | 0 | — | — | 0 | — | — |
| *21* | 100 | 0.14 ± 0.33 | -0.20 ± 0.26 | 100 | 0.54 ± 1.02 | -0.20 ± 0.26 |
| *22* | 100 | 1.32 ± 1.25 | 0.00 ± 0.00 | 100 | 13.02 ± 12.37 | 0.00 ± 0.00 |
| *23* | 100 | 0.99 ± 1.12 | 0.00 ± 0.00 | 100 | 9.83 ± 10.40 | 0.00 ± 0.00 |
| *24* | 100 | 0.25 ± 0.54 | -0.25 ± 0.36 | 100 | 3.72 ± 3.88 | -0.25 ± 0.36 |
| *25* | 100 | 1.74 ± 1.31 | 0.00 ± 0.00 | 100 | 17.38 ± 12.94 | 0.00 ± 0.00 |
| *26* | 100 | 0.02 ± 0.17 | -0.20 ± 0.31 | 100 | 0.44 ± 0.62 | -0.20 ± 0.31 |
| *27* | 98 | 0.10 ± 0.37 | -0.29 ± 0.40 | 98 | 2.28 ± 1.90 | -0.29 ± 0.40 |
| *28* | 100 | 1.67 ± 1.18 | 0.00 ± 0.00 | 100 | 16.93 ± 12.03 | 0.00 ± 0.00 |
| *29* | 100 | 0.03 ± 0.08 | -0.13 ± 0.26 | 100 | 0.31 ± 0.30 | -0.13 ± 0.26 |
| *30* | 100 | -0.06 ± 0.51 | -0.62 ± 0.41 | 100 | 2.63 ± 2.80 | -0.62 ± 0.41 |
| *31* | 100 | -0.03 ± 0.10 | -0.44 ± 0.40 | 100 | 0.24 ± 0.30 | -0.44 ± 0.40 |
| *32* | 96 | -0.22 ± 0.41 | -0.82 ± 0.25 | 96 | 0.91 ± 1.59 | -0.82 ± 0.25 |
| *33* | 100 | 0.01 ± 0.02 | -0.38 ± 0.33 | 100 | 0.12 ± 0.10 | -0.38 ± 0.33 |

Table 2: Relative (to LVE) performance of $\epsilon$-LVE with $\epsilon = 0.1$ and $\epsilon = 0.01$ (numbers are truncated).

Shenoy and Shafer [22] introduced the framework of valuation algebras, which states the basic algebraic requirements for efficient computation with valuations. More recently, Haenni [11] incorporated partially ordered preferences in the algebra to enable approximate computation. Fargier et al. [9] then extended the framework with a preference degree structure in order to capture the common algebraic structure of optimization problems based on a partial order. The algebra we develop in Section 3 can be partly casted in this framework.

The variable elimination algorithm we develop here is conceptually close to the message passing algorithm of Dubus et al. [7]. Their algorithm, however, does not handle uncertainty and target primarily the obtention of Pareto-efficient solutions for a specific class of multi-objective optimization problems.

There is a close relation between maximum a posteriori (MAP) inference in Bayesian networks and LIMIDs whose decision variables have no parents. In this sense, the algorithm of de Campos [4], which solves MAP by propagating Pareto efficient probability potentials in a join tree, relates to ours.

# 7 Conclusion

Solving limited memory influence diagrams is a very hard task. The complexity results presented here show that the problem is NP-complete even for diagrams with bounded treewidth and number of states per variable, and that obtaining provably good approximations in polynomial time is unlikely if the number of states is not small. Remarkably, as we show here, if the cardinalities of the variable domains are bounded by a constant, the problem does have a fully polynomial time approximation scheme.

Despite the theoretical hardness of the problem, we developed an algorithm that performed empirically well on a large set of randomly generated problems. The algorithm efficiency is based on the early removal of suboptimal solutions, which helps the algorithm to drastically reduce the search space. In the worst case, the algorithm runs in time exponential in both the width of the elimination ordering and the cardinality of decision variables.

In the experiments we conducted, the approximation did not result in a speed up of running time compared to the exact algorithm. This might be caused by the large constants produced by the boundedness assumptions, but might also be due to the ability of the exact algorithm in discarding many intermediate solutions. We note, however, that the simple existence of an efficient approximation shows that faster algorithms might exist, for instance, by allowing additive instead of multiplicative errors, or by coupling these ideas into a more sophisticated framework of propagation of functions.

Designing good heuristics for elimination orderings seems to be more complex than with standard variable elimination algorithms (e.g., for belief updating in Bayesian networks), because there is a second component, the cardinality of a set, that together with domain cardinalities we wish to minimize. In fact, some preliminary experimentation has shown that favoring set cardinality at expense of domain cardinality might be a good approach. Unlike standard variable elimination, given an elimination ordering and a LIMID, it does not seem to be possible to determine the true complexity of LVE in advance (i.e., prior to running the algorithm).

# Appendix A. Missing Proofs and Additional Results

This section contains long proofs that were left out of the main part to improve readability, and less central results used in some of the proofs. Some of the results in here are based on results obtained elsewhere and reproduced here (albeit with minor modifications) for completeness and ease of reading, but most are contributions of this paper. Results that are largely based on previous results contain a mention to the source; otherwise it is a new result.

The following two lemmas are used in the proof of Theorem 4 later on.

**Lemma 32.** *If $\alpha \geq -2$ is a real number and $i$ is a nonnegative integer then* $2^{\alpha} + 2^{-(i+3)} < 2^{\alpha+2^{-i}}$.

*Proof.* A similar result was shown by de Campos [4, Lemma 15]. Since $2^{\alpha} \geq 2^{-2}$, we have that $2^{\alpha} + 2^{-(i+3)} = 2^{\alpha} + 2^{-2} \cdot 2^{-i-1} \leq 2^{\alpha}(1 + 2^{-i-1})$, and it is sufficient to show that $1 + 2^{-i-1} < 2^{2^{-i}}$. From the Binomial Theorem we have that

$$(1 + 2^{-i-1})^{2^i} = \sum_{k=0}^{2^i} \binom{2^i}{k} (2^{-i-1})^k \,.$$

For $k = 0, \ldots, 2^i$, we have that

$$\binom{2^i}{k} = \frac{2^i(2^i - 1) \cdots (2^i - k + 1)}{k!} \leq (2^i)^k \,.$$

Hence,

$$(1 + 2^{-i-1})^{2^i} \leq \sum_{k=0}^{2^i} (2^i)^k (2^{-i-1})^k = \sum_{k=0}^{2^i} 2^{-k} \leq \sum_{k=0}^{\infty} 2^{-k} = 2 \,,$$

and therefore $1 + 2^{-i-1} < 2^{2^{-i}}$. □

**Lemma 33.** *If $0 \leq x \leq 1/2$ then $2^{x-1} + 2^{-x-1} \geq 2^{x^4}$.*

*Proof.* We obtain the result by approximating the functions on the left- and right-hand side of the inequalities by their truncated Taylor expansions $f(x)$ and $g(x)$, respectively, and then showing that $2^{x-1} + 2^{-x-1} \geq f(x) \geq g(x) \geq 2^{x^4}$. The $n$-th order Taylor expansion of the left-hand side around zero is given by

$$T_n(x) = 1 + \sum_{k=1}^{n/2} \frac{[\ln(2)]^{2k}}{(2k)!} x^{2k} \,.$$

Clearly, the series converges and hence $2^{x-1} + 2^{-x-1} = \lim_{n\to\infty} T_n(x)$. Moreover, for any $n$, the residual $R_n(x) = 2^{x-1} + 2^{-x-1} - T_n(x)$ is positive because the terms of the sum are all non negative. Thus,

$$f(x) = T_2(x) = 1 + \frac{[\ln(2)]^2}{2}x^2 \leq 2^{x-1} + 2^{-x-1}\,.$$

In a similar fashion, we apply the variable change $y = x^4$ on the right-hand side and obtain its Taylor expansion around zero, given by

$$\begin{aligned} T'_n(y) &= 1 + \sum_{k=1}^{n} \frac{[\ln(2)]^k}{k!} y^k \\ &= 1 + \sum_{k=1}^{n} \frac{[\ln(2)]^k}{k!} x^{4k}\,, \end{aligned}$$

which also converges and has positive residual. Hence,

$$\begin{aligned} 2^{x^4} &= \lim_{n\to\infty} T'_n(x) \\ &= 1 + x^4\ln(2) + x^2\ln(2)\left(\sum_{k=2}^{\infty} \frac{[\ln(2)]^{k-1}}{k!} x^{4k-2}\right) \\ &\leq 1 + x^4\ln(2) + x^2\ln(2)\left(\sum_{k=2}^{\infty} \frac{1}{2^{4k-1}}\right) \\ &= 1 + x^4\ln(2) + \frac{[\ln(2)]^2}{32}x^2 = g(x)\,. \end{aligned}$$

The inequality is obtained by noticing that $[\ln(2)]^{k-1}/k! < 1/2$, $x \leq 1/2 \leq \ln(2)$ and that the geometric series

$$\sum_{k=2}^{\infty} \frac{1}{2^{4k-1}} = \frac{1}{2^7}\sum_{k=0}^{\infty}\left(\frac{1}{2^4}\right)^k < \frac{1}{2^7}\sum_{k=0}^{\infty}\left(\frac{1}{2}\right)^k = \frac{1}{2^6} < \frac{\ln(2)}{32}\,.$$

Finally, since $x^2 \leq 1/4 < 15\ln(2)/32$ we have that

$$\begin{aligned} g(x) &= 1 + x^2\ln(2)\left(x^2 + \frac{\ln(2)}{32}\right) \\ &< 1 + x^2\ln(2)\left(\frac{15}{32}\ln(2) + \frac{\ln(2)}{32}\right) \\ &= 1 + \frac{[\ln(2)]^2}{2}x^2 = f(x)\,. \end{aligned}$$

Hence, $2^{x^4} \leq g(x) \leq f(x) \leq 2^{x-1} + 2^{-x-1}$ and the result holds. □

*Proof of Theorem 4.* Given a strategy $s$, deciding whether $E_s[\mathcal{L}] > k$ can be done in polynomial time according to Proposition 3.

Hardness is shown using a reduction from the *partition* problem, which is NP-complete [10] and can be stated as follows. *Given a set of n positive integers* $a_1, \ldots, a_n$*, is there a set* $\mathcal{I} \subset \mathcal{A} = \{1, \ldots, n\}$ *such that* $\sum_{i\in\mathcal{I}} a_i = \sum_{i\in\mathcal{A}\setminus\mathcal{I}} a_i$*?* We assume that $n > 3$.

Let $a = \frac{1}{2}\sum_{i\in\mathcal{A}} a_i$. An *even partition* is a subset $\mathcal{I} \subset \mathcal{A}$ that achieves $\sum_{i\in\mathcal{I}} a_i = a$. To solve *partition*, we consider the rescaled problem (dividing every element by $a$), so that $v_i = a_i/a \leq 2$ are the elements and we look for a partition such that $\sum_{i\in\mathcal{I}} v_i = 1$ (because $\sum_{i\in\mathcal{A}} v_i = 2$).

Consider the following LIMID with topology as in Figure 1. There are $n$ binary decision nodes labeled $D_1, \ldots, D_n$. Each decision $D_i$ can take on states $\boldsymbol{d}_1$ and $\boldsymbol{d}_2$. The chain of chance nodes has $n+1$ ternary variables $X_0, X_1, \ldots, X_n$ with states $\boldsymbol{x}$, $\boldsymbol{y}$, and $\boldsymbol{z}$. There is an arc from $X_n$ to the single value node $R$. For notational purposes, we specify a function $f$ over the domain $\{\boldsymbol{x}, \boldsymbol{y}, \boldsymbol{z}\}$ as a triple $(f(\boldsymbol{x}), f(\boldsymbol{y}), f(\boldsymbol{z}))$. The value node has an associated utility function $u_R = (0, 0, 1)$. For $i = 1, \ldots, n$, each chance node $X_i$ has an associated set of conditional probability mass functions given by

$$
\begin{aligned}
p_{X_i}^{\boldsymbol{d}_1,\boldsymbol{x}} &= (t_i, 0, 1-t_i), & p_{X_i}^{\boldsymbol{d}_2,\boldsymbol{x}} &= (1, 0, 0),\\
p_{X_i}^{\boldsymbol{d}_1,\boldsymbol{y}} &= (0, 1, 0), & p_{X_i}^{\boldsymbol{d}_2,\boldsymbol{y}} &= (0, t_i, 1-t_i),\\
p_{X_i}^{\boldsymbol{d}_1,\boldsymbol{z}} &= (0, 0, 1), & p_{X_i}^{\boldsymbol{d}_2,\boldsymbol{z}} &= (0, 0, 1),
\end{aligned}
$$

for $t_i \in [0,1]$ (we specify these variables later on). Note that $p_{X_i}^{D_i X_{i-1}}(\boldsymbol{w}) = 0$ for every $\boldsymbol{w} \in \Omega_{\mathtt{fa}_{X_i}}$ such that $\boldsymbol{w}^{\downarrow X_i} \neq \boldsymbol{w}^{\downarrow X_{i-1}}$ and $\boldsymbol{w}^{\downarrow X_i} \neq \boldsymbol{z}$. Finally, we define $p_{X_0} = (1/3, 1/3, 1/3)$.

Given a strategy $s = (\delta_{D_1}, \ldots, \delta_{D_n})$, let $\mathcal{I} \triangleq \{i : \delta_{D_i} = \boldsymbol{d}_1\}$ be the index set of policies in $s$ such that $\delta_{D_i}(\lambda) = \boldsymbol{d}_1$. We have that

$$
\begin{aligned}
E_s[\mathcal{L}] &= \sum_{\mathcal{C}\cup\mathcal{D}} \left( p_{X_0} \prod_{i=1}^{n} p_{X_i}^{D_i X_{i-1}} p_{D_i} \right) u_R \\
&= \sum_{X_n} \left( \sum_{\mathcal{C}\cup\mathcal{D}\setminus\{X_n\}} p_{X_0} \prod_{i=1}^{n} p_{X_i}^{D_i X_{i-1}} p_{D_i} \right) u_R \,.
\end{aligned}
$$

Let

$$
p_s \triangleq p_{X_0} \prod_{i=1}^{n} p_{X_i}^{D_i X_{i-1}} p_{D_i}
$$

and

$$
p_{X_n} \triangleq \sum_{\mathcal{C}\cup\mathcal{D}\setminus\{X_n\}} p_{X_0} \prod_{i=1}^{n} p_{X_i}^{D_i X_{i-1}} p_{D_i} = \sum_{\mathcal{C}\cup\mathcal{D}\setminus\{X_n\}} p_s \,.
$$

For $\boldsymbol{w} \in \Omega_{\mathcal{C}\cup\mathcal{D}}$ such that $\boldsymbol{w}^{\downarrow X_n} = \boldsymbol{x}$ (i.e., for $\boldsymbol{w} \in \boldsymbol{x}^{\uparrow\mathcal{C}\cup\mathcal{D}}$) it follows that $p_{X_n}^{D_n X_{n-1}}(\boldsymbol{w}^{\downarrow \mathtt{fa}_{X_n}}) \neq 0$ if and only if $\boldsymbol{w}^{\downarrow X_{n-1}} = \boldsymbol{x}$. But for $\boldsymbol{w}^{\downarrow X_{n-1}} = \boldsymbol{x}$ we have that $p_{X_{n-1}}^{D_{n-1} X_{n-2}}(\boldsymbol{w}^{\downarrow \mathtt{fa}_{X_{n-1}}}) \neq 0$ if and only if $\boldsymbol{w}^{\downarrow X_{n-2}} = \boldsymbol{x}$ and so recursively. Also, for any $i \in \{1, \ldots, n\}$, $p_{X_i}^{D_i X_{i-1}}(\boldsymbol{w}^{\downarrow \mathtt{fa}_{X_i}})$ equals $t_i$ if $i \in \mathcal{I}$ and 1 otherwise. Hence,

$$
p_s(\boldsymbol{w}) = \begin{cases} \frac{1}{3}\prod_{i\in\mathcal{I}} t_i, & \text{if } \boldsymbol{w}^{\downarrow X_i} = \boldsymbol{x} \text{ for } i = 1, \ldots, n-1 \\ 0, & \text{otherwise,} \end{cases}
$$

and

$$
p_{X_n}(\boldsymbol{x}) = \sum_{\boldsymbol{w}\in\boldsymbol{x}^{\uparrow\mathcal{C}\cup\mathcal{D}}} p_s(\boldsymbol{w}) = \frac{1}{3} \prod_{i\in\mathcal{I}}^{n} t_i \,.
$$

Likewise, it holds for $\boldsymbol{w} \in \boldsymbol{y}^{\uparrow \mathcal{C} \cup \mathcal{D}}$ that

$$p_s(\boldsymbol{w}) = \begin{cases} \frac{1}{3} \prod_{i \in \mathcal{A} \setminus \mathcal{I}} t_i, & \text{if } \boldsymbol{w}^{\downarrow X_i} = \boldsymbol{y} \text{ for } i = 1, \dots, n-1 \\ 0, & \text{otherwise,} \end{cases}$$

and therefore

$$p_{X_n}(\boldsymbol{x}) = \frac{1}{3} \prod_{i \in \mathcal{A} \setminus \mathcal{I}}^{n} t_i \,.$$

Since $p_{X_n}$ is a probability mass function on $X_n$, $p_{X_n}(\boldsymbol{z}) = 1 - p_{X_n}(\boldsymbol{x}) - p_{X_n}(\boldsymbol{y})$, and

$$\begin{aligned} \mathrm{E}_s[\mathcal{L}] &= \sum_{X_n} p_{X_n} u_R \\ &= 1 - p_{X_n}(\boldsymbol{x}) - p_{X_n}(\boldsymbol{y}) \\ &= 1 - \frac{1}{3} \prod_{i \in \mathcal{I}} t_i - \frac{1}{3} \prod_{i \in \mathcal{A} \setminus \mathcal{I}} t_i \,. \end{aligned}$$

Let us assume initially that $t_i = 2^{-v_i}$. The reduction from the original problem in this way is not polynomial, and we will use it only as an upper bound for the outcome of the reduction we obtain later. It is not difficult to see that $\mathrm{E}_s[\mathcal{L}]$ is a concave function of $v_1, \dots, v_n$ that achieves its maximum at $\sum_{i \in \mathcal{I}} v_i = \sum_{i \in \mathcal{A} \setminus \mathcal{I}} v_i = 1$. Since each strategy $s$ defines a partition of $\mathcal{A}$ and vice-versa, there is an even partition if and only if $\mathrm{MEU}[\mathcal{L}] = 1 - 1/3(1/2 + 1/2) = 2/3$.

We will now show a reduction that encodes the numbers $t_i$ in time and space polynomial in $b$, the number of bits used to encode the original problem. In fact, this is in close analogy with the final part of the proof of Theorem 10 in [4].

By setting $t_i$ to represent $2^{-v_i}$ with $6b+3$ bits of precision (rounding up if necessary), that is, by choosing $t_i$ so that $2^{-v_i} \le t_i < 2^{-v_i} + \epsilon_i$, where $0 \le \varepsilon_i < 2^{-(6b+3)}$, we have that $2^{-v_i} \le t_i < 2^{-v_i} + 2^{-(6b+3)}$, which by Lemma 32 (with $\alpha = -v_i \ge -2$ and $i = 6b$) implies $2^{-v_i} \le t_i < 2^{-v_i + 2^{-6b}}$.

Assume that an even partition $\mathcal{I}$ exists. Then[11]

$$\prod_{i \in \mathcal{I}} t_i < 2^{2^{-6b} n - \sum_{i \in \mathcal{I}} v_i} = 2^{-1 + 2^{-6b} n} \le 2^{-1 + 2^{-5b}} \,,$$
$$\prod_{i \in \mathcal{A} \setminus \mathcal{I}} t_i < 2^{2^{-6b} n - \sum_{i \in \mathcal{A} \setminus \mathcal{I}} v_i} = 2^{-1 + 2^{-6b} n} \le 2^{-1 + 2^{-5b}} \,,$$

and

$$\mathrm{MEU}[\mathcal{L}] > 1 - \frac{1}{3} \left( 2^{-1 + 2^{-5b}} + 2^{-1 + 2^{-5b}} \right) = 1 - \frac{2^{2^{-5b}}}{3}. \tag{6}$$

Let $r$ be equal to $2^{2^{-5b}}$ encoded with $5b+3$ bits of precision (and rounded up), that is, $2^{2^{-5b}} \le r < 2^{2^{-5b}} + 2^{-(5b+3)}$, which by Lemma 32 (with $\alpha = 2^{-5b} \ge -2$ and $i = 5b$) implies

$$2^{2^{-5b}} \le r < 2^{2^{-5b} + 2^{-5b}} = 2^{2^{1-5b}} < 2^{2^{-4b}} \,. \tag{7}$$

[11] Since the number of bits used to encode the partition problem must be greater than or equal to $n$, we have that $n/2^b \le n/b \le 1$, and hence $2^{-(j+1)b} n < 2^{-jb}$, for any $j > 0$.

The reduction is done by verifying whether $\text{MEU}[\mathcal{L}] > 1 - r/3$. We already know that an even partition has an associated strategy which obtains an expected utility greater than $1 - r/3$, because of Equality (6) and the fact that $r$ is rounded up. Let us consider the case where an even partition does not exist. We want to show that in this case $\text{MEU}[\mathcal{L}] \leq 1 - 2^{2^{-4b}}/3$, which by Inequality (7) implies $\text{MEU}[\mathcal{L}] < 1 - r/3$. Since there is not an even partition, any strategy induces a partition such that, for some integer $-a \leq c \leq a$ different from zero, we have that $\sum_{i \in \mathcal{I}} a_i = a - c$ and $\sum_{i \in \mathcal{A} \setminus \mathcal{I}} a_i = a + c$, because the original numbers $a_i$ are positive integers that add up to $2a$. It follows that

$$\prod_{i \in \mathcal{I}} t_i + \prod_{i \in \mathcal{A} \setminus \mathcal{I}} t_i = 2^{c/a-1} + 2^{-c/a-1} .$$

The right-hand side of the equality is a function on $c \in \{-a, \ldots, a\} \setminus \{0\}$, which is symmetric with respect to the y-axis (i.e., $f(c) = f(-c)$) and monotonically increasing for $c > 0$. Therefore, it obtains its minimum at $c = 1$. Hence,

$$\prod_{i \in \mathcal{I}} t_i + \prod_{i \in \mathcal{A} \setminus \mathcal{I}} t_i \geq 2^{1/a-1} + 2^{-1/a-1} .$$

Since $n > 3$ implies $a \geq 2$ (because the numbers $a_i$ are positive integers), we have by Lemma 33 that

$$2^{1/a-1} + 2^{-1/a-1} \geq 2^{1/a^4} .$$

Each number $a_i$ is encoded with at least $\log_2 a_i$ bits, and therefore $b \geq \log_2(a_1) + \cdots + \log_2(a_n) = \log_2(a_1 \cdots a_n)$. The latter is greater than or equal to $\log_2(a_1 + \cdots + a_n)$, and hence is also greater than $\log_2 a$. Thus, we have that $a \leq 2^b$, which implies $a^4 \leq 2^{4b}$ and therefore $1/a^4 \geq 2^{-4b}$ and $2^{1/a^4} \geq 2^{2^{-4b}}$. Hence,

$$2^{1/a-1} + 2^{-1/a-1} \geq 2^{2^{-4b}} .$$

Thus, if an even partition does not exist we have that

$$\text{MEU}[\mathcal{L}] = 1 - \frac{1}{3} \left( \prod_{i \in \mathcal{I}} t_i + \prod_{i \in \mathcal{A} \setminus \mathcal{I}} t_i \right) \leq 1 - \frac{2^{2^{-4b}}}{3} < 1 - r/3 .$$

To summarize, we have built a LIMID $\mathcal{L}$ in polynomial time since each $t_i$ was specified using $O(b)$ bits and there are $n$ functions $p_{X_i}^{D_i X_{i-1}}$, each encoding 18 numbers (which are either 1, 0 or $t_i$), and $2n + 2$ variables with bounded number of states. We have shown that there is a one-to-one correspondence between partitions of $\mathcal{A}$ in the original problem and strategies of $\mathcal{L}$, and that for a given rational $r = f(b)$ encoded with $O(b)$ bits the existence of an even partition is equivalent to $\text{MEU}[\mathcal{L}] > 1 - r/3$. ∎

The following lemma is used in the proof of Theorem 6.

**Lemma 34.** *For any $x \geq 1$ it follows that $x + 1/2 > 1/\ln(1 + 1/x)$.*

*Proof.* Adapted from Lemma 9 of [18]. Let $f(x) = \ln(1 + 1/x) - 1/(x + 1/2)$. Then

$$f'(x) = -\frac{1}{x^2 + x} + \frac{1}{x^2 + x + 1/4} ,$$

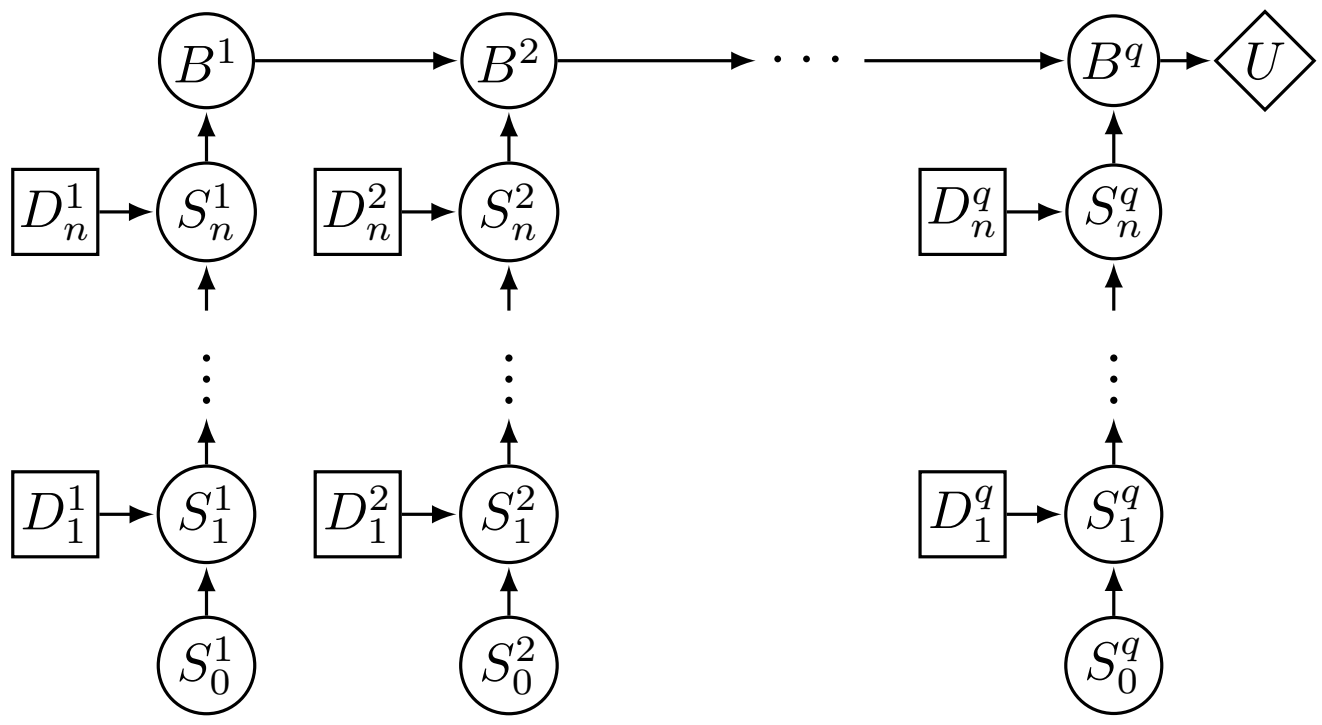

Figure 8: Graph structure of the LIMID used in the proof of Theorem 6.

which is strictly negative for $x \geq 1$ since $x^2+x < x^2+x+1/4$. Hence, $f(x)$ is a monotonically decreasing function for $x \geq 1$. Because $\lim_{x\to\infty} f(x) = 0$, $f(x)$ is strictly positive in $[1,\infty)$. Thus, the result follows from $\ln(1+1/x) > 1/(x+1/2)$, since $x \geq 1$. ☐

*Proof of Theorem 6.* We will show that for any fixed $0 < \gamma < 1$ the existence of a polynomial time $(2^{\theta\gamma}-1)$-approximation algorithm for solving a LIMID would imply the existence of a polynomial time algorithm for the CNF-SAT problem, which is known to be impossible unless P=NP [10]. A very similar reduction was used by Park and Darwiche [18, Theorem 8] to show an analogous inapproximability result for maximum a posteriori inference in Bayesian networks. Notice that for any $0 < \epsilon < 2^\theta - 1$ there is $\gamma < 1$ such that $\epsilon = 2^{\theta\gamma}-1$, hence the existence of an $\epsilon$-approximation algorithm implies the existence of a $(2^{\theta\gamma}-1)$-approximation, and it suffices for the desired result to show that the latter cannot be true (unless P=NP).

A clause is a disjunction of literals, each literal being either a boolean variable or its negation. We say that a clause is satisfied if, given an assignment of truth values to its variables, at least one of the literals evaluates to 1. Thus, we can decide if a truth-value assignment satisfies a clause in time linear in the number of variables. The CNF-SAT problem is defined as follows. *Given a set of clauses $C_1,\ldots,C_m$ over (subsets of) boolean variables $X_1,\ldots,X_n$, is there an assignment of truth values to the variables that satisfies all the clauses?*

For a positive integer $q$ that we specify later on, consider the LIMID obtained as follows (the topology is depicted in Figure 8). For each boolean variable $X_i$ we add $q$ binary decision variables $D_i^1,\ldots,D_i^q$ and $q$ chance variables $S_i^1,\ldots,S_i^q$ with domain $\{0,1,\ldots,m\}$. Additionally, there are $q$ clause selector variables $S_0^1,\ldots,S_0^q$ taking values on $\{1,2,\ldots,m\}$, $q$ binary variables $B^1,\ldots,B^q$, and a value node $U$ with $B^q$ as parent. As illustrated in Figure 8, the LIMID consists of $q$ replicas of a polytree-shaped diagram over variables $D_1^j,\ldots,D_n^j,S_0^j,\ldots,S_n^j,B^j$, and the probability mass functions for the variables $B^1,\ldots,B^q$ are chosen so as to make the expected utility equal the product of the expected utilities of each replica. In any of the replicas (i.e., for $j \in \{1,\ldots,q\}$), a variable $D_i^j$ $(i=1,\ldots,n)$ represents an assignment of truth value for $X_i$ and has no parents. The selector variables $S_0^j$ represent the choice of a clause to pro-cess, that is, $S_0^j = k$ denotes clause $C_k$ is being "processed", and by summing out $S_0^j$ we process all clauses. Each variable $S_i^j$, for $i = 1, \ldots, n$ and $j = 1, \ldots, q$, has $D_i^j$ and $S_{i-1}^j$ as parents. The variables $B^j$ have $S_n^j$ and, if $j > 1$, $B^{j-1}$ as parents. For all $j$, we assign uniform probabilities to $S_0^j$, that is, $p_{S_0^j} \triangleq 1/m$. For $j = 1, \ldots, q$, we set the probabilities associated to variables $S_1^j, \ldots, S_n^j$ so that if $C_k$ is the clause selected by $S_0^j$ then $S_i^j$ is set to zero if $C_k$ is satisfied by $D_i$ but not by any of $D_1, \ldots, D_{i-1}$, and $S_i^j = S_{i-1}^j$ otherwise. Formally, for $\boldsymbol{x} \in \Omega_{\{S_i^j, D_i^j, S_{i-1}^j\}}$ we have that

$$p_{S_i^j}^{D_i^j S_{i-1}^j}(\boldsymbol{x}) \triangleq \begin{cases} 1, & \text{if } \boldsymbol{x}^{\downarrow S_i^j} = \boldsymbol{x}^{\downarrow S_{i-1}^j} = 0\,; \\ 1, & \text{if } \boldsymbol{x}^{\downarrow S_i^j} = 0 \text{ and } \boldsymbol{x}^{\downarrow S_{i-1}^j} = k \geq 1 \text{ and } X_i = \boldsymbol{x}^{\downarrow D_i} \text{ satisfies } C_k\,; \\ 1, & \text{if } \boldsymbol{x}^{\downarrow S_i^j} = \boldsymbol{x}^{\downarrow S_{i-1}^j} = k \geq 1 \text{ and } X_i = \boldsymbol{x}^{\downarrow D_i} \text{ does not satisfy } C_k\,; \\ 0, & \text{otherwise.} \end{cases}$$

Notice that for $S_1^j$ the first case never occurs since $S_0^j$ takes values on $\{1, \ldots, m\}$. For any joint state configuration $\boldsymbol{x}$ of $S_0^j, \ldots, S_n^j, D_1^j, \ldots, D_n^j$ such that $\boldsymbol{x}^{\downarrow S_0^j} = k \in \{1, \ldots, m\}$ (i.e., clause $C_k$ is being processed) and $\boldsymbol{x}^{\downarrow S_n^j} = 0$, it follows that

$$\left( p_{S_0^j} \prod_{i=1}^{n} p_{S_i^j}^{D_i^j S_{i-1}^j} p_{D_i^j} \right)(\boldsymbol{x})$$

equals $1/m$ only if for some $0 < i \leq n$ clause $C_k$ is satisfied by $X_i = \boldsymbol{x}^{\downarrow D_i}$ but not by any of $X_1 = \boldsymbol{x}^{\downarrow D_1}, \ldots, X_{i-1} = \boldsymbol{x}^{\downarrow D_{i-1}}$, variables $S_1^j, \ldots, S_{i-1}^j$ all assume value $k$ (i.e., $\boldsymbol{x}^{\downarrow S_1^j} = \cdots = \boldsymbol{x}^{\downarrow S_{i-1}^j} = k$), and $\boldsymbol{x}^{\downarrow S_i^j} = \cdots = \boldsymbol{x}^{\downarrow S_n^j} = 0$. Otherwise, it equals 0. Hence, for any (partial) strategy $s^j = (\delta_{D_1^j}, \ldots, \delta_{D_n^j})$ we have for $\boldsymbol{x} = 0$ that

$$p_{S_n^j}^{s^j}(\boldsymbol{x}) \triangleq \left( \sum_{\substack{S_0^j, \ldots, S_{n-1}^j \\ D_1^j, \ldots, D_n^j}} p_{S_0^j} \prod_{i=1}^{n} p_{S_i^j}^{D_i^j S_{i-1}^j} p_{D_i^j} \right)(\boldsymbol{x}) = \frac{SAT(s^j)}{m}\,,$$

where $SAT(s^j)$ denotes the number of clauses satisfied by the truth-value assignment of $X_1, \ldots, X_n$ according to $s^j$. Each variable $B^j$ is associated to a function $p_{B^j}^{S_n^j B^{j-1}}$ such that for $\boldsymbol{x} \in \Omega_{\mathtt{fa}_{B^j}}$,

$$p_{B^j}^{S_n^j B^{j-1}}(\boldsymbol{x}) = \begin{cases} 1, & \text{if } \boldsymbol{x}^{\downarrow B^j} = \boldsymbol{x}^{\downarrow B^{j-1}} \text{ and } \boldsymbol{x}^{\downarrow S_n^j} = 0\,; \\ 1, & \text{if } \boldsymbol{x}^{\downarrow B^j} = 0 \text{ and } \boldsymbol{x}^{\downarrow S_n^j} \neq 0\,; \\ 0, & \text{otherwise;} \end{cases}$$

where for $B^1$ we assume $\boldsymbol{x}^{\downarrow B^0} = 1$. Hence, we have for any joint state configuration $\boldsymbol{x}$ of $B^1, \ldots, B^q, S_n^1, \ldots, S_n^q$ that

$$\left( \prod_{j=1}^{q} p_{B^j}^{S_n^j B^{j-1}} \right)(\boldsymbol{x}) = \begin{cases} 1, & \text{if } \boldsymbol{x}^{\downarrow B^1} = \cdots = \boldsymbol{x}^{\downarrow B^q} = 1 \text{ and } \boldsymbol{x}^{\downarrow S_n^1} = \cdots = \boldsymbol{x}^{\downarrow S_n^q} = 0; \\ 1, & \text{if } \boldsymbol{x}^{\downarrow B^1} = \cdots = \boldsymbol{x}^{\downarrow B^q} = 0 \text{ and } \boldsymbol{x}^{\downarrow S_n^1} \neq 0; \\ 0, & \text{otherwise.} \end{cases}$$

Finally, we set the utility function $u$ associated to $U$ to return 1 if $B^q = 1$ and 0 otherwise. In this way, $\left(u \prod_{j=1}^{q} p_{B^j}^{S_n^j B^{j-1}}\right)(\boldsymbol{x})$ equals 1 if $\boldsymbol{x}^{\downarrow B^1} = \cdots = \boldsymbol{x}^{\downarrow B^q} = 1$ and $\boldsymbol{x}^{\downarrow S_n^1} = \cdots = \boldsymbol{x}^{\downarrow S_n^q} = 0$ and zero otherwise. Thus, for any strategy $s = (s^1, \ldots, s^q)$, where $s^j = \delta_{D_1^j}, \ldots, \delta_{D_n^j}$, it follows that

$$\begin{aligned}
\mathrm{E}_s[\mathcal{L}] &= \sum_{\mathcal{C}\cup\mathcal{D}} u \prod_{j=1}^{q} p_{B^j}^{S_n^j B^{j-1}} p_{S_0^j} \prod_{i=1}^{n} p_{S_i^j}^{D_i^j S_{i-1}^j} p_{D_i^j} \\
&= \sum_{\substack{B^1,\ldots,B^q \\ S_n^1,\ldots,S_n^q}} u \prod_{j=1}^{q} p_{B^j}^{S_n^j B^{j-1}} \sum_{\substack{S_0^j,\ldots,S_{n-1}^j \\ D_1^j,\ldots,D_n^j}} p_{S_0^j} \prod_{i=1}^{n} p_{S_i^j}^{D_i^j S_{i-1}^j} p_{D_i^j} \\
&= \sum_{\substack{B^1,\ldots,B^q \\ S_n^1,\ldots,S_n^q}} u \prod_{j=1}^{q} p_{B^j}^{S_n^j B^{j-1}} p_{S_n^j}^{s^j} \\
&= \prod_{j=1}^{q} p_{S_n^j}^{s^j}(0) = \frac{1}{m^q} \prod_{j=1}^{q} SAT(s^j)\,.
\end{aligned}$$

If the instance of CNF-SAT problem is satisfiable then there is an optimum strategy $s$ such that $SAT(s^j) = m$ for all $j$, and $\mathrm{MEU}[\mathcal{L}] = 1$. On the other hand, if the instance is not satisfiable, we have for all $j$ and strategy $s$ that $SAT(s^j) \leq m-1$, and hence $\mathrm{MEU}[\mathcal{L}] \leq (m-1)^q/m^q$. For some given $0 < \gamma < 1$, let $q$ be a positive integer chosen so that $1/2^{\theta^\gamma} > m^q/(m+1)^q$. We show later on that $q$ can be obtained from a polynomial on the input. If the CNF-SAT instance is satisfiable, a $(2^{\theta^\gamma} - 1)$-approximation algorithm for $\mathrm{MEU}[\mathcal{L}]$ returns a value $\mathrm{E}_s[\mathcal{L}]$ such that

$$\mathrm{E}_s[\mathcal{L}] \geq \frac{\mathrm{MEU}[\mathcal{L}]}{2^{\theta^\gamma}} > \left(\frac{m}{m+1}\right)^q > \left(\frac{m-1}{m}\right)^q ,$$

where the rightmost strict inequality follows from $m/(m+1) > (m-1)/m$. On the other hand, if the CNF-SAT instance is not satisfiable, the approximation returns

$$\mathrm{E}_s[\mathcal{L}] \leq \mathrm{MEU}[\mathcal{L}] \leq \left(\frac{m-1}{m}\right)^q .$$

Hence, we can use a $(2^{\theta^\gamma} - 1)$-approximation algorithm to solve CNF-SAT by checking whether its output $\mathrm{E}[\mathcal{L}] > (m-1)^q/m^q$. Since $q$ and $m$ are positive integers, the test bound $(m-1)^q/m^q$ can be obtained in polynomnial time.

It remains to show that the reduction is polynomial in the input. The LIMID contains $q(2n+2)+1$ variables, each requiring the specification of at most $2(m+1)^2$ numbers in $\{0, 1/m, 1\}$. So $\theta$, the number of numerical parameters in $\mathcal{L}$, is polynomially bounded by $q(m+1)^2(4n+4)+2$. Therefore, it suffices to show that $q$ is a polynomial on $m$ and $n$. By definition, $q$ obeys

$$\left(1 + \frac{1}{m}\right)^q > 2^{[q(m+1)^2(4n+4)+2]^\gamma} ,$$

which is equivalent to

$$q \ln\left(1+\frac{1}{m}\right) > q^{\gamma}[(m+1)^2(4n+4)+2]^{\gamma}\ln 2$$
$$\Leftrightarrow q^{1-\gamma} > \frac{[(m+1)^2(4n+4)+2]^{\gamma}}{\ln\left(1+\frac{1}{m}\right)}\ln 2$$
$$\Leftrightarrow q > \left(\frac{[(m+1)^2(4n+4)+2]^{\gamma}}{\ln\left(1+\frac{1}{m}\right)}\ln 2\right)^{\frac{1}{1-\gamma}}$$

Since by Lemma 34, $m + 1/2 > 1/\ln(1 + 1/m)$ and $2 > \ln(2)$, it suffices to choose $q$ such that

$$q > \left((2m+1)[(m+1)^2(4n+4)+2]^{\gamma}\right)^{\frac{1}{1-\gamma}} .$$

In other words, $q$ is polynomially bounded by $m^{\frac{2\gamma+1}{1-\gamma}}4n^{\frac{\gamma}{1-\gamma}}$. Therefore, if MEU[$\mathcal{L}$] can be approximated in polynomial time with an error no greater than $2^{\theta\gamma}$ then we can solve CNF-SAT in polynomial time. □

We now formally prove the correctness of the LVE algorithm. We start by showing that max distributes over set marginalization and set combination:

**Lemma 35.** *(Distributivity of maximality). If $\Psi_x \subset \Phi_x$ and $\Psi_y \subset \Phi_y$ are two finite sets of ordered valuations and $z \subseteq x$, the following holds.*

*(i)* $\max(\Psi_x \otimes \max(\Psi_y)) = \max(\Psi_x \otimes \Psi_y)$*;*

*(ii)* $\max(\max(\Psi_x)^{\downarrow z}) = \max(\Psi_x^{\downarrow z})$.

*Proof.* Part (i) has been shown by Fargier et al. [9, Lemma 1(iv)]. We use a similar proof to show that part (ii) also holds. First, we show that $\max(\Psi_x^{\downarrow z}) \subseteq \max(\max(\Psi_x)^{\downarrow z})$. Assume, to show a contradiction, that there is an element $\phi_x^{\downarrow z} \in \max(\Psi_x^{\downarrow z})$, where $\phi_x \in \Psi_x$, which is not an element of $\max(\max(\Psi_x)^{\downarrow z})$. By definition of $\max(\Psi_x)$, there is $\psi_x \in \max(\Psi_x)$ such that $\phi_x \leq \psi_x$. Hence, (A5) implies $\phi_x^{\downarrow z} \leq \psi_x^{\downarrow z}$, and because $\psi_x^{\downarrow z} \in \Psi_x^{\downarrow z}$ it follows that $\phi_x^{\downarrow z} = \psi_x^{\downarrow z}$, and therefore $\phi_x^{\downarrow z} \in \max(\Psi_x)^{\downarrow z}$. Since $\phi_x^{\downarrow z} \notin \max(\max(\Psi_x)^{\downarrow z})$ there is $\phi_z \in \max(\max(\Psi_x)^{\downarrow z})$ such that $\phi_x^{\downarrow z} \leq \phi_z$. But this contradicts our initial assumption since $\phi_z \in \Psi_x^{\downarrow z}$.

Let us now show that $\max(\Psi_x^{\downarrow z}) \supseteq \max(\max(\Psi_x)^{\downarrow z})$. Assume by contradiction that there is $\psi_z \in \max(\max(\Psi_x)^{\downarrow z}) \setminus \max(\Psi_x^{\downarrow z})$. Since $\psi_z \in \Psi_x^{\downarrow z}$, there is $\phi_z \in \max(\Psi_x^{\downarrow z})$ such that $\psi_z \leq \phi_z$. But we have shown that $\max(\Psi_x^{\downarrow z}) \subseteq \max(\max(\Psi_x)^{\downarrow z})$, hence $\psi_z = \phi_z$ and $\psi_z \in \max(\Psi_x^{\downarrow z})$, a contradiction. □

At any iteration $i$ of the propagation step, the combination of all sets in the current pool of sets $\mathcal{V}_i$ produces the set of maximal valuations of the initial factorization marginalized to $X_{i+1}, \ldots, X_n$:

**Lemma 36.** *For $i \in \{0, 1, \ldots, n\}$, it follows that*

$$\max\left(\left[\bigotimes_{\Psi\in\mathcal{V}_0}\Psi\right]^{-\{X_1,\ldots,X_i\}}\right) = \max\left(\bigotimes_{\Psi\in\mathcal{V}_i}\Psi\right),$$

*where for each $i$, $\mathcal{V}_i$ is the collection of sets of valuations generated at the $i$-th iteration of the propagation step of LVE.*

*Proof.* By induction on $i$. The basis ($i = 0$) follows trivially.

Assume the result holds at $i$, that is,

$$\max\left(\left[\bigotimes_{\Psi\in\mathcal{V}_0}\Psi\right]^{-\{X_1,\ldots,X_i\}}\right) = \max\left(\bigotimes_{\Psi\in\mathcal{V}_i}\Psi\right) .$$

By eliminating $X_{i+1}$ from both sides and then applying the max operation we get to

$$\max\left(\left[\max\left(\left[\bigotimes_{\Psi\in\mathcal{V}_0}\Psi\right]^{-\{X_1,\ldots,X_i\}}\right)\right]^{-X_{i+1}}\right) = \max\left(\left[\max\left(\bigotimes_{\Psi\in\mathcal{V}_i}\Psi\right)\right]^{-X_{i+1}}\right) .$$

Applying Lemma 35(ii) to both sides and (A2) to the left-hand side yields

$$\begin{aligned}
\max\left(\left[\bigotimes_{\Psi\in\mathcal{V}_0}\Psi\right]^{-\{X_1,\ldots,X_{i+1}\}}\right) &= \max\left(\left[\bigotimes_{\Psi\in\mathcal{V}_i}\Psi\right]^{-X_{i+1}}\right)\\
&= \max\left(\left[\bigotimes_{\Psi\in\mathcal{V}_i\setminus\mathcal{B}_{i+1}}\Psi\right]\otimes\left[\bigotimes_{\Psi\in\mathcal{B}_{i+1}}\Psi\right]^{-X_{i+1}}\right)\\
&= \max\left(\left[\bigotimes_{\Psi\in\mathcal{V}_i\setminus\mathcal{B}_{i+1}}\Psi\right]\otimes\max\left(\left[\bigotimes_{\Psi\in\mathcal{B}_{i+1}}\Psi\right]^{-X_{i+1}}\right)\right)\\
&= \max\left(\left[\bigotimes_{\Psi\in\mathcal{V}_i\setminus\mathcal{B}_{i+1}}\Psi\right]\otimes\Psi_i\right)\\
&= \max\left(\bigotimes_{\Psi\in\mathcal{V}_{i+1}}\Psi\right) ,
\end{aligned}$$

where the passage from the first to the second identity follows from element-wise application of (A1) and Lemma 11, the third follows from the second by Lemma 35(i), and the last two follow from the definitions of $\Psi_i$ and $\mathcal{V}_{i+1}$, respectively. □

We are now able to show the correctness of the algorithm in solving LIMIDs exactly.

*Proof of Theorem 19.* The algorithm returns the utility part of a valuation $(p, u)$ in $\max\left(\bigotimes_{\Psi\in\mathcal{V}_n}\Psi\right)$, which, by Lemma 36 for $i = n$, equals $\max\left(\left[\bigotimes_{\Psi\in\mathcal{V}_0}\Psi\right]^{\downarrow\emptyset}\right)$. By definition of $\mathcal{V}_0$, any valuation $\phi$ in $\left(\bigotimes_{\Psi\in\mathcal{V}_0}\Psi\right)$ factorizes as in (5). Also, there is exactly one valuation $\phi\in\left(\bigotimes_{\Psi\in\mathcal{V}_0}\Psi\right)$ for each strategy in $\Delta$. Hence, by Proposition 12, the set $\left(\bigotimes_{\Psi\in\mathcal{V}_0}\Psi\right)^{\downarrow\emptyset}$ contains a pair $(1, \mathrm{E}_s[\mathcal{L}])$ for every strategy $s$ inducing a distinct expected utility. Moreover, since functions with empty scope correspond to numbers, the relation $\leq$ specifies a total ordering over the valuations in $\left(\bigotimes_{\Psi\in\mathcal{V}_0}\Psi\right)^{\downarrow\emptyset}$, which implies a single maximal element. Let $s^*$ be a strategy associated to $(p,u)$. Since $(p,u) \in \max\left(\left[\bigotimes_{\Psi\in\mathcal{V}_0}\Psi\right]^{\downarrow\emptyset}\right)$, it follows from maximality that $\mathrm{E}_{s^*}[\mathcal{L}] \geq \mathrm{E}_s[\mathcal{L}]$ for all $s$, and hence $u = \mathrm{MEU}[\mathcal{L}]$. $\square$

*Proof of Proposition 22.* Consider a variable $D$ in $\mathcal{L}$ and let $D_1,\ldots,D_m$ be the corresponding decision variables and $X_1,\ldots,X_m$ the corresponding chance variables in $\mathcal{L}'$. Also, let

$$\mathcal{P} \triangleq \left\{\sum_y \prod_{i=1}^m p_{X_i}^{\mathtt{pa}_{X_i}} p_{\boldsymbol{d}_i} : (p_{\boldsymbol{d}_1},\ldots,p_{\boldsymbol{d}_m}) \in \mathcal{P}_{D_1}\times\cdots\times\mathcal{P}_{D_m}\right\},$$

where $y = \{D_1, X_1, \ldots, D_{m-1}, X_{m-1}, D_m\}$, and, for each $D_i$, $p_{\boldsymbol{d}_i}$ denotes the probability mass function that assigns all mass to $\boldsymbol{d}_i \in \Omega_D$ (hence each set $\mathcal{P}_{D_i}$ contains $|\Omega_D|$ functions, and the set $\mathcal{P}_{D_1}\times\cdots\times\mathcal{P}_{D_m}$ has $|\Omega_D|^m$ elements). It suffices for the result to show that $\mathcal{P}$ is equal to $\mathcal{P}_D$. The functions $p \in \mathcal{P}$ have scope equal to $\{X_m\}\cup\mathtt{pa}_D$, and domain $\Omega_{\{X_m\}\cup\mathtt{pa}_D} = \Omega_{\mathtt{fa}_D}$. Consider $p\in\mathcal{P}$, and let $w \triangleq y\cup\{X_m\}\cup\mathtt{pa}_D$. For $\boldsymbol{x}\in\Omega_{\mathtt{fa}_D}$, let $1\leq j\leq m$ be such that $\boldsymbol{x}^{\downarrow\mathtt{pa}_D} = \boldsymbol{\pi}_j$. Thus,

$$p(\boldsymbol{x}) = \sum_{\boldsymbol{y}\in\boldsymbol{x}^{\uparrow w}} p_{X_j}^{\mathtt{pa}_{X_j}}(\boldsymbol{y}^{\downarrow\mathtt{fa}_{X_j}}) p_{\boldsymbol{d}_j}(\boldsymbol{y}^{\downarrow D_j}) \prod_{i\neq j} p_{X_i}^{\mathtt{pa}_{X_i}}(\boldsymbol{y}^{\downarrow\mathtt{fa}_{X_i}}) p_{\boldsymbol{d}_i}(\boldsymbol{y}^{\downarrow D_i}).$$

For all $i\neq j$, the values $p_{X_i}^{\mathtt{pa}_{X_i}}(\boldsymbol{y}^{\downarrow\mathtt{fa}_{X_j}})$ do not depend on the realization of $\boldsymbol{y}^{\downarrow D_i}$. Hence,

$$p(\boldsymbol{x}) = \sum_{\boldsymbol{y}\in\boldsymbol{x}^{\uparrow w\setminus z}} p_{X_j}^{\mathtt{pa}_{X_j}}(\boldsymbol{y}^{\downarrow\mathtt{fa}_{X_j}}) p_{\boldsymbol{d}_j}(\boldsymbol{y}^{\downarrow D_j}) \prod_{i\neq j} p_{X_i}^{\mathtt{pa}_{X_i}}(\boldsymbol{y}^{\downarrow\mathtt{fa}_{X_i}}) \left(\sum_z \prod_{i\neq j} p_{\boldsymbol{d}_i}\right),$$

where $z = \{D_1,\ldots,D_m\}\setminus\{D_j\}$, and the term inside the parentheses is the sum-marginal of $\prod_{i\neq j} p_{\boldsymbol{d}_i}$ over $z$. Because each $p_{\boldsymbol{d}_i}$ is a probability mass function on $D_i$ this term equals one and we have that

$$p(\boldsymbol{x}) = \sum_{\boldsymbol{y}\in\boldsymbol{x}^{\uparrow w\setminus z}} p_{\boldsymbol{d}_j}(\boldsymbol{y}^{\downarrow D_j}) \prod_{i=1}^m p_{X_i}^{\mathtt{pa}_{X_i}}(\boldsymbol{y}^{\downarrow\mathtt{fa}_{X_i}}).$$

Now, for all $i\neq j$, the values $p_{X_i}^{\mathtt{pa}_{X_i}}(\boldsymbol{y}^{\downarrow\mathtt{fa}_{X_j}})$ equal one if $\boldsymbol{y}^{\downarrow X_i} = \boldsymbol{y}^{\downarrow X_{i-1}}$ and zero otherwise. In addition, the values $p_{X_j}^{\mathtt{pa}_{X_j}}(\boldsymbol{y}^{\downarrow\mathtt{fa}_{X_j}})$ equal one if $\boldsymbol{y}^{\downarrow X_j} = \boldsymbol{y}^{\downarrow X_{j-1}} = \boldsymbol{y}^{\downarrow D_j}$ and zero otherwise. Hence, the product $\prod_{i=1}^m p_{X_i}^{\mathtt{pa}_{X_i}}(\boldsymbol{y}^{\downarrow\mathtt{fa}_{X_i}})$ differs from zero only if $\boldsymbol{y}^{\downarrow X_1} = \cdots = \boldsymbol{y}^{\downarrow X_m} = \boldsymbol{y}^{\downarrow D_j}$, in which case it equals one. Since $p_{\boldsymbol{d}_j}(\boldsymbol{y}^{\downarrow D_j})$ equals one if $\boldsymbol{y}^{\downarrow D_j} = \boldsymbol{d}_j$ and zero otherwise, we have that

$$p(\boldsymbol{x}) = \begin{cases} 1, & \text{if } \boldsymbol{x}^{\downarrow X_m} = \boldsymbol{d}_j\,; \\ 0, & \text{otherwise.}\end{cases}$$

Notice that for each $\boldsymbol{\pi}_j \in \Omega_{\mathtt{pa}_D}$ there is exactly one $\boldsymbol{d}\in\Omega_D$ such that $p(\boldsymbol{d},\boldsymbol{\pi}_j) = 1$, and hence $\mathcal{P}$ is the set of degenerate conditional mass functions on $\Omega_D$. Since the set $\mathcal{P}_D$ contains a function $p_D^{\mathtt{pa}_D}$ for every possible combination of degenerate mass functions on $\Omega_D$ (one mass function for each $\boldsymbol{\pi}_j\in\Omega_{\mathtt{pa}_D}$), it follows that for each $p$ there is $p_D^{\mathtt{pa}_D}$ such that $p = p_D^{\mathtt{pa}_D}$. Thus, $\mathcal{P}\subseteq\mathcal{P}_D$.

Consider a function $p_D^{\mathsf{pa}_D} \in \mathcal{P}_D$ and its associated policy $\delta_D \in \Delta_D$. For $i = 1, \ldots, m$ let $p_{\boldsymbol{d}_i}$ be the function from $\mathcal{P}_{D_i}$ assigning all mass to $\boldsymbol{d}_i = \delta(\boldsymbol{\pi}_i)$. Also, let $p$ be a function in $\mathcal{P}$ such that $\sum_y \prod_{i=1}^m p_{X_i}^{\mathsf{pa}_{X_i}} p_{\boldsymbol{d}_i}$. Then, for $\boldsymbol{d} \in \Omega_D$ and $\boldsymbol{\pi}_i \in \Omega_{\mathsf{pa}_D}$, $p_D^{\mathsf{pa}_D}(\boldsymbol{d}, \boldsymbol{\pi}_i) = p_{\boldsymbol{d}_i}(\boldsymbol{d}) = p(\boldsymbol{d}, \boldsymbol{\pi}_i)$. Hence, for any function $p_D^{\mathsf{pa}_D}$ there is $p \in \mathcal{P}$ such that $p_D^{\mathsf{pa}_D} = p$, and $\mathcal{P}_D \subseteq \mathcal{P}$. □

The following inequalities are required for the proof of Theorem 31.

**Lemma 37.** *For any nonnegative integer $k$ and $0 \leq x \leq 1$,*

$$1 + 2x \geq \left(1 + \frac{x}{k}\right)^k .$$

*Proof.* From the Binomial Theorem of Elementary Algebra we have that

$$\left(1 + \frac{x}{k}\right)^k = \sum_{i=0}^{k} \binom{k}{i} \frac{x^i}{k^i} \leq \sum_{i=0}^{k} \frac{x^i}{i!} \leq \sum_{i=0}^{\infty} \frac{x^i}{i!} = e^x ,$$

because

$$\binom{k}{i} = \frac{k(k-1)\cdots(k-i+1)}{i!} \leq \frac{k^i}{i!}$$

for $i = 0, \ldots, k$. Thus, it suffices to show that $e^x \leq 1 + 2x$, which is true if and only if $x \leq \ln(1 + 2x)$. Let $f(x) = \ln(1 + 2x) - x$. Then

$$f'(x) = \frac{2}{1 + 2x} - 1 \begin{cases} > 0, & \text{if } x < 1/2, \\ = 0, & \text{if } x = 1/2, \\ < 0, & \text{if } x > 1/2, \end{cases}$$

and therefore $f(x)$ monotonically increases from 0 to 1/2 and monotonically decreases from 1/2 to 1. Since $f(0) = 0$ and $f(1) = \ln(3) - 1 > 0$, it follows that $f \geq 0$ in $[0, 1]$ and hence $x \leq \ln(1 + 2x)$ for $x \in [0, 1]$. □

**Lemma 38.** *For any $x \geq 0$,*

$$\ln(1 + x) \geq \frac{x}{1 + x} .$$

*Proof.* Let $f(x) = \ln(1 + x) - x/(1 + x)$. Then $f$ is a monotonically increasing function for $x \geq 0$ because

$$f'(x) = \frac{x}{(1 + x)^2} .$$

Since $f(0) = 0$, $\ln(1+x) - x/(1+x) \geq 0$ for all $x \geq 0$ and the result follows. □